\documentclass{article}

\usepackage{xcolor}

\PassOptionsToPackage{hypertexnames=false}{hyperref}

\usepackage[preprint]{corl_2026} % preprint version

\usepackage{algorithm}
\usepackage{algpseudocode}
\usepackage{amsmath}
\usepackage{amssymb}
\usepackage{bm}
\usepackage{booktabs}
\usepackage{colortbl}
\usepackage{graphicx}
\usepackage{wrapfig}
\usepackage{mathtools}
\usepackage{xparse}
\usepackage{xspace}
\usepackage{float}
\usepackage{capt-of}
\usepackage{tikz}
\usepackage{subcaption}
\usepackage{etoc}
\usepackage{titletoc}
\usepackage{tabularx}
\usepackage{array}
\usepackage{makecell}
\usepackage{graphicx}
\usetikzlibrary{fadings}

\usepackage[capitalise]{cleveref}
\crefname{equation}{Eq.}{Eqs.}
\crefname{figure}{Fig.}{Figs.}
\crefname{tabular}{Tab.}{Tabs.}
\crefname{section}{Sec.}{Secs.}
\crefname{algocf}{Alg.}{Algs.}
\crefname{algocfline}{Line}{Lines}
\crefname{lem}{Lemma}{Lemmata}

\newcommand{\proj}{\operatorname{proj}}

\newcommand{\regtext}[1]{\mathrm{\textnormal{#1}}}

\newcommand{\mc}[1]{\mathcal{#1}}
\newcommand{\lbl}[1]{_{\regtext{#1}}}

\newcommand{\vc}[1]{\mathbf{#1}} % vectorize stuff
\newcommand{\mat}[1]{{\begin{bmatrix} #1 \end{bmatrix}}}

\newcommand{\eye}{\vc{I}}

\newcommand{\inv}{^{-1}}
\newcommand{\invbrac}[1]{\left[ {#1} \right]\inv}

\newcommand{\norm}[1]{\left\Vert#1\right\Vert}

\DeclareMathOperator*{\argmin}{arg\,min}

\newcommand{\rodrigues}{\mc{R}}
\newcommand{\subproblem}[1]{\method{SP#1}}
\DeclareMathOperator{\normalize}{unit}

\newcommand{\set}[1]{\left\{#1\right\}}

\newcommand{\R}{\mathbb{R}}

\newcommand{\SO}{\regtext{SO}}
\newcommand{\SE}{\regtext{SE}}

\newcommand{\method}[1]{\textnormal{#1}\xspace}
\newcommand{\ourmethod}{\method{WARP}}

\newcommand{\sewmimic}{\method{SEW-Mimic}}
\newcommand{\sew}{\method{SEW}}

\newcommand{\makeframe}{\method{MakeFrame}}

\newcommand{\config}{\vc{q}}

\newcommand{\rotmat}{\vc{R}}
\newcommand{\transvec}{\vc{p}}
\newcommand{\pose}{\vc{T}}
\newcommand{\dataset}{\mathcal{D}}
\newcommand{\humandata}{\mathcal{H}}

\newcommand{\robotdata}{\mathcal{B}}

\newcommand{\frm}[1]{^{#1}}
\newcommand{\frms}[2]{^{#1 \leftarrow #2}}

\newcommand{\robot}{\regtext{rb}}
\newcommand{\human}{\regtext{hm}}
\newcommand{\object}{\regtext{obj}}

\newcommand{\leftside}{\regtext{L}}
\newcommand{\rightside}{\regtext{R}}
\newcommand{\lside}[1][]{_{\leftside\if\relax\detokenize{#1}\relax\else,\,#1\fi}}
\newcommand{\rside}[1][]{_{\rightside\if\relax\detokenize{#1}\relax\else,\,#1\fi}}

\newcommand{\foft}{(t)}
\newcommand{\arms}{_{(\lside,\rside)}}

\newcommand{\shoulder}{\vc{s}}
\newcommand{\elbow}{\vc{e}}
\newcommand{\wrist}{\vc{w}}
\newcommand{\tool}{\vc{t}}

\newcommand{\handorientation}{\vc{H}}

\newcommand{\upperarm}{\vc{u}}
\newcommand{\lowerarm}{\vc{l}}

\newcommand{\length}{\ell}
\newcommand{\keypoints}{K}
\newcommand{\robotkeypoints}[1][]{\keypoints_{\regtext{rb}\if\relax\detokenize{#1}\relax\else,\,#1\fi}}

\def\papermode{draft}
\def\draftmode{draft}

\def\Snospace~{\S{}}

\definecolor[named]{TODORed}{HTML}{FF2828}
\definecolor[named]{TLDRViolet}{HTML}{7B1FA2}
\definecolor[named]{MissingCyan}{HTML}{00838F}
\definecolor[named]{PendingOrange}{HTML}{EF6C00}
\definecolor[named]{cBlue}{HTML}{1565C0}
\definecolor[named]{cOrange}{HTML}{EF6C00}
\definecolor[named]{cPink}{HTML}{AD1457}
\definecolor[named]{cYellow}{HTML}{AF8F00}
\definecolor[named]{cGreen}{HTML}{2E7D32}
\definecolor[named]{cGray}{HTML}{616161}
\definecolor[named]{cNavy}{HTML}{000080}
\definecolor[named]{cPurple}{RGB}{130,80,180}

\ifx\papermode\draftmode
    \newcommand{\todo}[1]{{\color{TODORed}\textbf{#1}}}
    \newcommand{\TODO}[1]{{\color{TODORed}\textbf{[TODO]} #1}}
    \newcommand{\tldr}[1]{\medskip\noindent{\color{TLDRViolet}\textbf{[TL;DR]} #1}}
    
    \newcommand{\placeholder}[1]{{\color{PendingOrange}\textbf{[placeholder]} #1}}
    \NewDocumentCommand{\makecomment}{m m m o}{%
      \noindent{\color{#2}[\textbf{#1}] #3}%
      \IfValueT{#4}{{\color{cGray}\space\textit{#4}}}%
    }
\else
    \newcommand{\todo}[1]{}
    \newcommand{\TODO}[1]{}
    \newcommand{\tldr}[1]{}
    
    \newcommand{\placeholder}[1]{}
    \NewDocumentCommand{\makecomment}{m m m o}{}
\fi

\NewDocumentCommand{\zhenyang}{m o}{\makecomment{ZC}{cYellow}{#1}[#2]}

\NewDocumentCommand{\kong}{m o}{\makecomment{CK}{cPink}{#1}[#2]}

\NewDocumentCommand{\chuye}{m o}{\makecomment{YE}{cPurple}{#1}[#2]}

\NewDocumentCommand{\atian}{m o}{\makecomment{AT}{cGreen}{#1}[#2]}

\NewDocumentCommand{\sidd}{m o}{\makecomment{SK}{cOrange}{#1}[#2]}

\NewDocumentCommand{\danfei}{m o}{\makecomment{DX}{cBlue}{#1}[#2]}

\NewDocumentCommand{\lawrence}{m o}{\makecomment{Lawrence}{cNavy}{#1}[#2]}
\NewDocumentCommand{\den}{m o}{\makecomment{Dennis}{teal}{#1}[#2]}
\NewDocumentCommand{\yuanshao}{m o}{\makecomment{Yuanshao}{cNavy}{#1}[#2]}

\title{WARP: Whole-Body Retargeting for Learning from Offline Human Demonstrations}
\author{
  Zhenyang Chen$^{*}$,
  Chuizheng Kong$^{*}$,
  Chuye Zhang$^{*}$, \\
  Yuanshao Yang,
  Lawrence Y. Zhu, \\
  Shreyas Kousik, and
  Danfei Xu\\
  $^{*}$ Equal contribution\\
  Georgia Institute of Technology, Atlanta, Georgia 30332
}

\begin{document}
\maketitle

\vspace{-10pt}
\begin{abstract}
% === 05/25 === %
% lawrence merge with finialized flow
Direct transfer from human demonstration to learnable robot action is a crucial step towards scalable whole-body mobile manipulation. While human data scales better than mobile teleoperation, it requires overcoming significant embodiment gaps. Existing retargeting methods yield imprecise or inconsistent solutions,
% \zc{also not faithfully generate wholebody motion}
causing action multi-modality that prevents supervised policies from reliably converging. We present Whole-body-Aware Retargeting from human Pose (WARP), an offline pipeline that explicitly models embodiment differences to extract precise, unique whole-body actions. WARP leverages a closed-form Shoulder-Elbow-Wrist (SEW) geometric solver for exact end-effector tracking while preserving whole-body structural intent.
Paired with lazy mobile-base control, it extracts accurate, consistent robot trajectories. %By resolving action multi-modality, WARP enables the reliable training of a hierarchy-masked flow policy.
Evaluations show WARP provides highly reliable data for open-loop real-world replay. To our knowledge, WARP is the first framework to achieve zero-shot whole-body mobile manipulation directly from offline human demonstrations, eliminating the need for human-in-the-loop teleoperation action data. More details on https://warp-retargeting.github.io/
\end{abstract}

\vspace{-10pt}
% \nolinenumbers
\begin{figure}[H]
  \centering
  \includegraphics[width=1.0\linewidth]{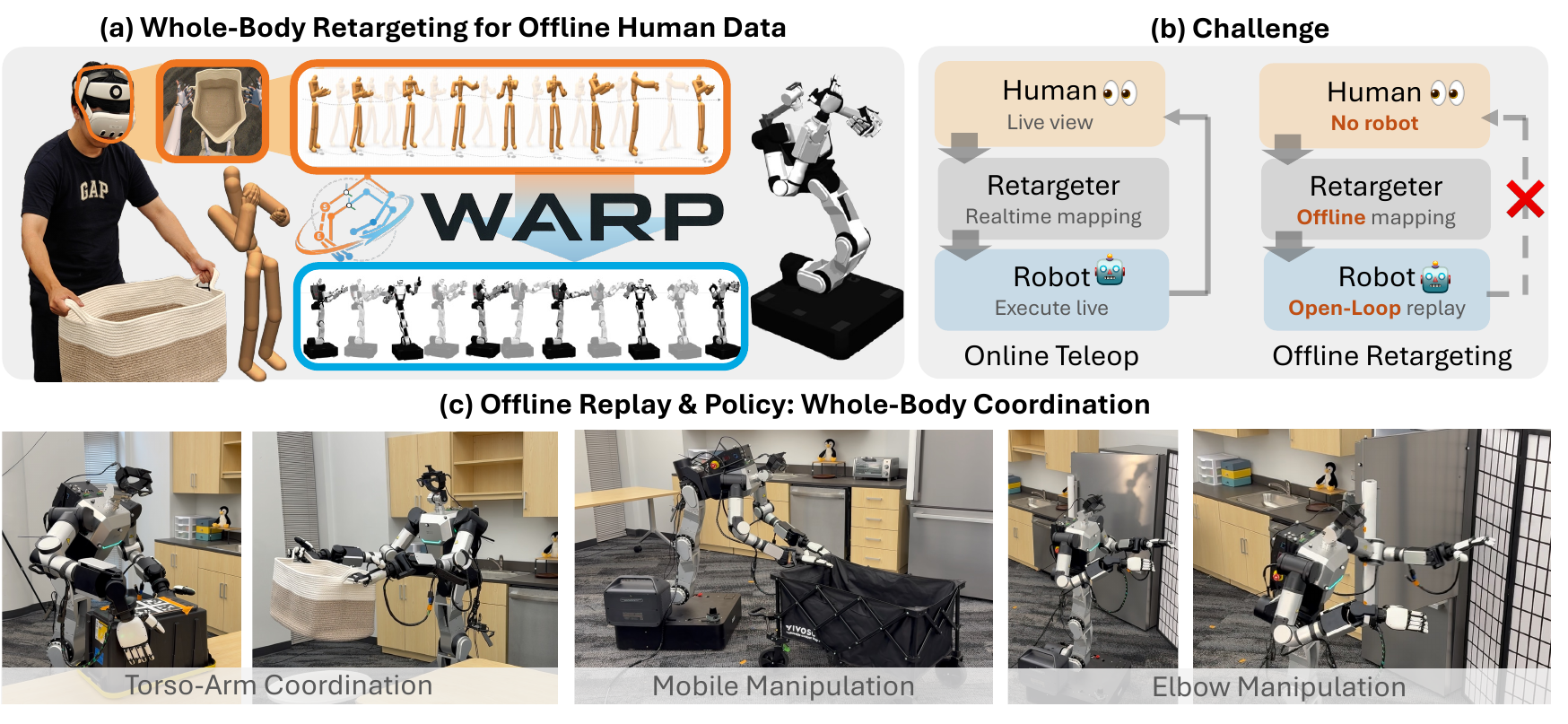}
  \vspace{-10pt}
    \caption{\small \textbf{Whole-body-Aware Retargeting from human Pose (WARP).}
    \textbf{(a)} We collect human manipulation data \emph{offline}
    using VR devices, and WARP retargets this motion into \emph{whole-body
    robot actions}, producing human-like trajectories directly usable for policy training.
    \textbf{(b)} The central difficulty of the offline setting is the absence
    of online human correction to close the embodiment gap. With no human in
    the loop to absorb mismatches, retargeting errors become far less
    tolerable, and small errors cause replay failure. \textbf{(c)} WARP retargets whole-body human motions, creating versatile whole-body coordination skills for manipulation.}
    \vspace{+5pt}
  \label{fig:teaser}
  \vspace{-1em}
\end{figure}
% \linenumbers

\section{Introduction}
\label{sec:introduction}
% ===== 05/26 ====%
% ZY: merge Danfei's draft with previous draft and Lawrence comment

% --- M: Motivation and the core challenge ---
Learning whole-body mobile manipulation policies directly from offline human demonstrations is an attractive path toward scale: compared to teleoperation data, human data is cheap to
record, needs no robot hardware in the loop, and captures unconstrained
whole-body behavior in natural environments.
The central difficulty is that a
human demonstration is not, by itself, a robot demonstration. Human motion
carries task-relevant structure that extends well beyond the hands: how the
torso is oriented, how the elbows route around obstacles, when the body shifts
to extend reach. However, a retargeter that merely places the robot end-effector near the
human hand discards the exact whole-body information that whole-body
manipulation depends on. To make human data usable, retargeting must convert
human motion into robot actions that are both \emph{physically replayable} on
the robot and \emph{statistically learnable} by a policy.

These requirements are far stricter in the offline setting than in online teleoperation. During teleoperation, a human operator closes the loop in real time: if the robot wrist drifts, misses contact, or selects an undesirable inverse-kinematics branch, the operator can compensate by changing their motion~\cite{fu2024mobilealoha, cheng2024opentelevision, xu2026hommi, araujo2025gmr}. Offline policy learning has no such corrective mechanism. The retargeted trajectory \emph{is} the supervision. This exposes a fundamental tension in whole-body retargeting: the robot must satisfy hard task constraints, especially end-effector poses, while also preserving the demonstrated human body configuration, including torso orientation, shoulder placement, elbow routing, and base motion. Prior offline methods have largely overlooked this tension~\cite{sundaresan2025homerlearninginthewildmobile, liu2025egozero, guzey2025aina}. %Common retargeters typically resolve it with weighted objectives~\cite{mink, araujo2025gmr}.
This creates two failure modes. First, the solution may be \emph{imprecise}: improving whole-body similarity can degrade end-effector tracking, while prioritizing the end effector can discard the torso, elbow, and base structure that made the manipulation feasible. Either case turns retargeting error into supervision error. Second, the solution may be \emph{inconsistent}: because humanoids are highly redundant, nearly identical human poses can map to different configurations depending on solver initialization or local tradeoffs. Similar observations would be paired with divergent actions, which impedes learning.

We address these issues with \textbf{WARP} (\cref{fig:teaser}), an offline human-to-robot retargeting and policy-learning framework for whole-body mobile manipulation. %WARP is built around two principles. First, retargeting should be \emph{consistent}: each human pose should map to a unique robot configuration, so the policy sees a single-mode supervisory signal. Second, retargeting should preserve the body structure that manipulation depends on: torso orientation, shoulder and elbow configuration, wrist pose, and base-body coordination, rather than reducing each demonstration to an end-effector trajectory. 
The key mechanism is the Shoulder-Elbow-Wrist (SEW) representation~\cite{kong2026sew}. Instead of directly optimizing over redundant joint angles, SEW represents the arm through its shoulder, elbow, wrist, and hand-frame geometry (Fig.~\ref{fig:main_method}). This converts retargeting from a weighted optimization problem into a set of \emph{closed-form geometric subproblems}. The wrist parameter constrains precise task-space poses (palm and fingers), while the elbow parameter preserves the human arm configuration within the remaining feasible set. Each non-degenerate human pose maps to a \emph{unique robot configuration}. WARP further extends the retargeting algorithm to a full system with constrained alignment for dexterous hands, a task-aware mechanism that selects which body parts to constrain, lazy lower-body control for smooth base-body coordination, and a hierarchical policy that efficiently learns whole-body actions. 

% Across retargeting and policy-learning evaluations, WARP produces precise and consistent robot trajectories, improves policy performance over end-effector-only and optimization-based retargeting baselines~\cite{mink, ben2025homie, araujo2025gmr, kong2026sew}, enables open-loop replay from offline human data, and transfers across multiple humanoid embodiments. To our knowledge, WARP is the first system to train a whole-body mobile manipulation policy directly from offline human demonstrations, without human-in-the-loop teleoperation.

% % --- R: Results highlights (ADDED — advisor's sketch stopped at K2) ---
% We evaluate WARP at two levels. At the retargeting level, on standard
% retargeting benchmarks WARP reduces whole-body tracking error by
% \textbf{[TODO: XX]\%} over prior baselines~\cite{mink, ben2025homie, araujo2025gmr, kong2026sew}
% while producing consistent solutions across initializations. At the policy
% level, identical policies trained on WARP-retargeted data outperform those
% trained on end-effector-only and optimization-based retargeting by
% \textbf{[TODO: metric / XX\%]}. These properties let offline human data be
% replayed open-loop on the robot and used directly as supervision: to our
% knowledge, this is the first whole-body mobile manipulation policy trained from
% offline human demonstrations alone, with zero-shot transfer on a tractable
% subset of real-world tasks. We further show that WARP applies across multiple
% humanoid embodiments.

We evaluate WARP at both the retargeting and policy-learning levels. On standard retargeting benchmarks, WARP reduces whole-body tracking error by 99.34\% over prior baselines~\cite{mink, ben2025homie, araujo2025gmr, kong2026sew} while producing consistent solutions across initializations. In downstream policy learning, identical policies trained on WARP-retargeted data outperform those trained on end-effector-only and optimization-based retargeting by 35\%. These results show that WARP produces robot trajectories that are precise, consistent, and learnable: they can be replayed open-loop on the robot, used directly as supervision, and transferred across multiple humanoid embodiments. To our knowledge, WARP is the first system to train a whole-body mobile manipulation policy directly from offline human demonstrations, without human-in-the-loop teleoperation, achieving zero-shot transfer on a tractable subset of real-world tasks.

\section{Related Work}
\label{sec:related-work}
% \subsection{Learning Mobile Manipulation from Demonstrations}
% === Outline ===
% 1.Talk about the behavior cloning pipeline, from data to policy learning and deployment.

\textbf{Behavior Cloning for Mobile Manipulation.}
Behavior cloning (BC) is the dominant paradigm for visuomotor robot learning:
collect expert demonstrations, fit a policy, and deploy it on the target robot.
These demonstrations are overwhelmingly gathered by teleoperation---
leader--follower rigs (\method{GELLO}, \method{Mobile~ALOHA}), VR systems
(\method{Open-TeleVision}, \method{Bunny-VisionPro}), and hand-held grippers
(\method{UMI}, \method{DexCap}) \citep{wu2024gello, fu2024mobilealoha,
cheng2024opentelevision, ding2024bunnyvisionpro, chi2024umi, wang2024dexcap}---
and feed expressive policies (\method{ACT}, \method{Diffusion~Policy}) and
vision--language--action models (\method{OpenVLA}, \method{$\pi_0$})
\citep{zhao2023learning, chi2023diffusionpolicy, kim2024openvla, black2024pi0}.
But BC couples data collection to deployment: because demonstrations are tied to
the embodiment they were collected on, every new robot needs fresh data. This
coupling is most costly for mobile manipulation, whose mobile, whole-body
platforms make teleoperation far more hardware- and effort-intensive. Recent
systems make strong progress on long-horizon
household tasks but remain teleoperation-driven \citep{xiong2024adaptive,
fu2024mobilealoha, jiang2025behavior, li2025clone, ben2025homie,
sundaresan2025homerlearninginthewildmobile, yang2025mobipi}: \method{Mobile~ALOHA}
enables low-cost bimanual whole-body collection, while \method{BEHAVIOR Robot
Suite}, \method{CLONE}, and \method{HOMIE} improve control for mobile or humanoid
data \citep{jiang2025behavior, li2025clone, ben2025homie}. Such systems improve
demonstration quality but stay locked to embodiment-specific teleoperation
hardware. We instead replace robot teleoperation with scalable human
demonstrations, using a retargeting system that closes the human--robot embodiment gap in the
kinematic action.

\textbf{Human to Robot Motion Retargeting.}
Retargeting converts human skeleton keypoints into robot joint trajectories.
Vision-based teleoperation systems (\method{AnyTeleop}, \method{Open-TeleVision})
map tracked upper-body motion to robots via embodiment-specific retargeting or
inverse kinematics, trading generality against geometric fidelity
\citep{qin2023anyteleop, cheng2024opentelevision}. At the humanoid scale,
retargeting is increasingly coupled with learned dynamics: \method{HOMIE} and
\method{CLONE} target locomotion-aware whole-body execution, while \method{GMR}
shows retargeting quality strongly affects downstream motion-tracking
\citep{ben2025homie, li2025clone, araujo2025gmr}. Our setting is quasi-static
and manipulation-centric, so we align most closely with \method{SEW-Mimic}, a
closed-form shoulder--elbow--wrist solver for upper-body humanoid teleoperation
\citep{kong2026sew}; its fast, precise upper-body alignment suits learning
mobile manipulation from human demonstrations, where closing the kinematic gap
matters more than general locomotion.

\section{Method}
\label{sec:method}
% ==== May 26 ====%
% --- Method, first paragraph ---
Learning from offline human demonstrations requires precise, feasible robot action data from retargeting.
% The retargeted trajectory is the policy's only supervision when learning from offline human demonstrations, and
Therefore, a useful retargeter must
satisfy two requirements: \emph{precision}---reproducing the human's whole-body
posture, accurately---and \emph{consistency}---mapping identical human
poses to a single robot configuration so the policy sees a unimodal target.
We meet both with \textbf{WARP}, a geometry-based retargeter paired with a policy
trained on its output. Its core, \method{c-SEW}, solves
retargeting in closed form over the Shoulder--Elbow--Wrist
representation~\citep{kong2026sew}, producing a \emph{unique} configuration that
exactly matches the human wrist and preserves whole-body pose similarity. A \emph{lazy lower-body controller} extend this
guarantee across the full task, and a \emph{hierarchy-masked flow policy}
(\cref{app:policy}) learns whole-body mobile manipulation directly from the
resulting demonstrations.

\noindent\textbf{Offline retargeting for policy learning.} \label{sec:problem-formulation}
Our goal is to convert offline human demonstrations into whole-body robot actions usable as
supervision for a behavior-cloning policy, with no operator in the
loop. 
Given a human whole-body trajectory $\tau^h=\{o_t,\humandata\}_{t=1}^T$---obervation $o_t$ and kinematic skeleton $\humandata$ information, captured by a MoCap or VR device---and a robot whose kinematic $\robotdata$ differs from the human's, an offline retargeter $Ret$
is a per-timestep map from $\humandata\foft$ to a robot action $q^r_t=Ret(\humandata\foft)$. The retargeted trajectory $\{q^r_t\}_{t=1}^{T}$ serves as both the
proprioceptive state and the supervision for behavior cloning: the policy
$\pi_\theta$ maps the environment observation $o_t$ and a state history
$q^r_{t-L:t}$ to the next action chunk $q^r_{t+1:t+H}$, trained by
$\min_{\theta}\,\mathbb{E}_{t,\,\tau^h}\,
\mathcal{L}\!\left(\pi_\theta(o_t, q^r_{t-L:t}),\; q^r_{t+1:t+H}\right)$.

\textbf{SEW representation for humanoid kinematics.}
The Shoulder--Elbow--Wrist (SEW) representation describes each arm by the geometric skeleton
$\humandata = (\shoulder,\elbow,\wrist,\handorientation, \tool)$, where $\shoulder,\elbow,\wrist, \tool \in\R^3$ are the shoulder, elbow, wrist positions, and palm offset and $\handorientation\in\SO(3)$ is the palm-center orientation, all expressed in the upper-body-centric frame $\pose\frm{\human}$ constructed from \cref{alg:make_frame}. Its key abstraction is to separate arm shape from embodiment scale. The human arm configuration is represented by the scale-invariant limb directions
$
\upperarm = \normalize(\elbow-\shoulder),
\lowerarm = \normalize(\wrist-\elbow).
$

For a humanoid arm with a spherical shoulder, pin-joint elbow, and spherical wrist, these SEW constraints admit a closed-form inverse-kinematic solution with at most one valid joint configuration~\citep{kong2026sew}. Thus, once a target robot skeleton
$(\shoulder\frm{\robot},\elbow\frm{\robot},\wrist\frm{\robot},\handorientation\frm{\robot})$
is specified, the upper-body joint action can be recovered deterministically:
$
\config\frm{\robot}
=
\sew\left(
\shoulder\frm{\robot},
\elbow\frm{\robot},
\wrist\frm{\robot},
\handorientation\frm{\robot}
\right).
$
The solution consistency of SEW provides a unique supervision signal and benefit policy learning.

However, vanilla SEW-solver~\citep{kong2026sew} aligns limb directions rather than the palm contact point. Define robot geometry as 
$\ell_{OS},\ell_{SE},\ell_{EW},\transvec\lbl{WT}$, read once from the URDF. Here $O,S,E,W,T$ denote the upper-body origin, shoulder, elbow, wrist, and palm/tool point, respectively. $\ell_{OS}$ is the origin-to-shoulder offset, $\ell_{SE}$ is the upper-arm length, $\ell_{EW}$ is the forearm length, and $\transvec\lbl{WT}\in\R^3$ is the fixed wrist-to-tool offset expressed in the hand frame. Since the human and robot have different link lengths, copying $\upperarm$, $\lowerarm$, and $\handorientation$ generally produces a systematic palm displacement:
$
\hat{\tool}\frm{\robot}
=
\shoulder\frm{\robot}
+ \length_{SE}\upperarm
+ \length_{EW}\lowerarm
+ \handorientation\frm{\human}\transvec\lbl{WT}, \
\hat{\tool}\frm{\robot}
\neq \tool.
$
This is acceptable for online teleoperation, where a human can compensate, but not for offline policy learning, where the retargeted action is the supervision. WARP therefore uses SEW as the deterministic IK backbone, but replaces direction-only matching with constrained skeleton alignment: the robot palm is forced to coincide with the demonstrated palm, while the remaining geometric freedom is used to preserve the human SEW structure.

\textbf{WARP's formulation.}
WARP departs from prior retargeters along two design axes. First, it constructs a target robot SEW skeleton
$(\shoulder\frm{\robot}, \elbow\frm{\robot}, \wrist\frm{\robot}, \handorientation\frm{\robot})$
before solving for joint angles, rather than optimizing directly in the redundant joint space $\config$. This decouples cross-embodiment skeleton alignment from joint-angle IK and lets the final robot action be recovered by the closed-form SEW solver. Second, WARP enforces palm matching as a hard constraint, rather than trading off end-effector accuracy and pose similarity through weighted objectives.

Denote the robot geometry defined previously by
$\robotdata=(\ell_{OS}, \ell_{SE}, \ell_{EW}, \transvec\lbl{WT})$. 
Where For each arm, WARP computes
\begin{align}
\bigl(
\pose\frms{\human}{\robot},
\shoulder\frm{\robot},
\elbow\frm{\robot},
\wrist\frm{\robot},
\handorientation\frm{\robot}
\bigr)\arms
=
\ourmethod\bigl(\humandata\arms,\robotdata\bigr),
\label{eq:warp-formulation}
\end{align}
satisfying
$
\bigl(
\wrist\frm{\robot}
+
\handorientation\frm{\robot}\transvec\lbl{WT}
\bigr)\arms
=
\bigl(
\invbrac{\pose\frms{\human}{\robot}}\tool
\bigr)\arms .
$
The constraint states that the robot palm must coincide exactly with the human palm in the aligned robot upper-body frame. The remaining degrees of freedom are resolved analytically to preserve the human SEW structure, as described next.
\ourmethod solves for the desired robot SEW representation in two stage: 1) adaptive offset that solves for the optimal robot torso placement, which accounts for major link length difference, 2) per-arm palm alignment that matches human and robot palm pose while solving for robot elbow nullspace use human elbow angle.

\begin{figure}[t]
  \centering
  \includegraphics[width=1.0\linewidth]{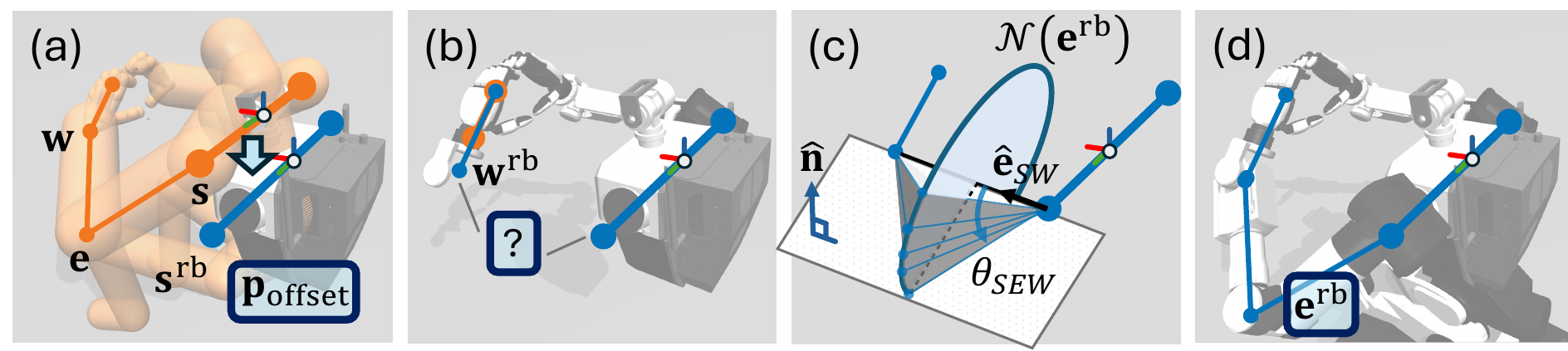}
  \vspace{-1em}
  \caption{ \small 
  Offline retargeting with \ourmethod. (a) Finding the optimal robot torso placement using Adaptive Offset (b) After aligning robot palm to human, robot wrist position can be solved (c) Prioritizing EEF alignment leaves the elbow configuration underconstrained. (d) Given fixed robot wrist and shoulder, we identify the elbow nullspace. Using \method{stereo-sew}~\cite{elias_redundancy_2024}, we find a unique plane intersecting the nullspace circle. (d) The solved robot SEW representation.
  }
  \vspace{-1em}
  \label{fig:main_method}
\end{figure}
% \linenumbers

% \subsection{Constrained Human and Robot Alignment}
% \label{sec:csew}

% \paragraph{Adaptive offset.}

\noindent\textbf{Adaptive offset.}
We first measure how far the robot palm drifts from the human palm when robot aligns its links in human arm directions $\upperarm\frm{\human}$, $\lowerarm\frm{\human}$.
Define the SEW forward kinematics for the robot palm:
$
    \hat{\tool}\frm{\robot}
    = \shoulder\frm{\robot}
      + \upperarm\frm{\human}\,\length_{SE}
      + \lowerarm\frm{\human}\,\length_{EW}
      + \handorientation\frm{\human}\,\transvec\lbl{WT},
    \label{eq:sew-fk}
$
where $\shoulder\frm{\robot}$ is the robot shoulder in the upper-body frame and $\transvec\lbl{WT} \in \R^3$ is the robot wrist-to-palm offset with $\norm{\transvec\lbl{WT}} = \length_{WT}$, read from the URDF.
The human palm $\tool$ is given directly from the body-centric frame extraction (\cref{sec:problem-formulation}).
The two-arm centroid offset that nullifies the net palm displacement is
\begin{equation}
    \transvec\lbl{offset}
    = \tfrac{1}{2}\!\bigl(\tool\lside + \tool\rside\bigr)
    - \tfrac{1}{2}\!\bigl(\hat{\tool}\frm{\robot}\lside + \hat{\tool}\frm{\robot}\rside\bigr).
    \label{eq:adaptive-offset}
\end{equation}
Shifting the robot upper-body origin by $\transvec\lbl{offset}$ as shown in \Cref{fig:main_method} (a) aligns the palm centroids exactly: $\pose\frms{\human}{\robot} \leftarrow (\transvec\lbl{offset}, \eye_3)$

\noindent\textbf{Per-arm palm alignment.}
With $\transvec\lbl{offset}$ applied, each arm is solved independently.
Hand orientation transfers directly, $\handorientation\frm{\robot} = \handorientation\frm{\human}$, and the desired robot wrist position is recovered from the human palm target $\tool$, transformed to robot frame using $\invbrac{\pose\frms{\human}{\robot}}$:
$
    \wrist\frm{\robot}
    = \invbrac{\pose\frms{\human}{\robot}}\tool - \handorientation\frm{\robot}\,\transvec\lbl{WT} + \transvec\lbl{offset}.
$
% which guarantees $
%     \wrist\frm{\robot} + \handorientation\frm{\robot}\,\transvec\lbl{WT}
%      = \invbrac{\pose\frms{\human}{\robot}}\tool$ by construction, enforcing the hard constraint of \cref{eq:cross-embodiment}.

It remains to find the robot elbow $\elbow\frm{\robot}$ consistent with $\shoulder\frm{\robot}$, $\wrist\frm{\robot}$, and link lengths $\length_{SE}$, $\length_{EW}$ as shown in \Cref{fig:main_method} (b).
We preserve the human elbow configuration by extracting the SEW elbow angle $\psi$ which allows us to transfer the unique elbow half-plane normal parametrization $\hat{\vc{n}}$ from the human skeleton to robot frame via the stereographic forward and inverse kinematics of \method{Stereo-sew}~\citep{elias_redundancy_2024}:
$
    \psi
    \;\leftarrow\; \method{Stereo-sew.FwdKin}\!\bigl(\shoulder, \elbow, \wrist\bigr)
$,
$
    \hat{\vc{n}}
    \; \leftarrow\; \method{Stereo-sew.InvKin}\!\bigl( \shoulder\frm{\robot}, \wrist\frm{\robot}, \psi \bigr).
$

Let $\hat{\vc{e}}_{SW} = \normalize(\wrist\frm{\robot} - \shoulder\frm{\robot})$.
We apply \subproblem{3} from \cite{elias2025ik} to find the rotation $\theta_{SEW}$ about $\hat{\vc{n}}$ that places the tip of the rotated upper-arm vector at distance $\length_{EW}$ from the desired wrist:
$
    \theta_{SEW}
    = \subproblem{3}\!\bigl(
        \length_{SE}\,\hat{\vc{e}}_{SW},\;
        \wrist\frm{\robot} - \shoulder\frm{\robot},\;
        \hat{\vc{n}},\;
        \length_{EW}
      \bigr).
$
Here, $\subproblem{3}$ solves for the direction of the robot upper arm $\length_{SE}$. 
Among the at most two solutions we select $\theta_{SEW} > 0$, the branch consistent with the human half-plane encoded by $\hat{\vc{n}}$.
The robot elbow position is then
\begin{equation}
    \elbow\frm{\robot}
    = \shoulder\frm{\robot}
      + \rodrigues\!\bigl(\hat{\vc{n}},\,\theta_{SEW}\bigr)\,
        \bigl(\length_{SE}\,\hat{\vc{e}}_{SW}\bigr),
    \label{eq:elbow-robot}
\end{equation}
where $\rodrigues(\hat{\vc{n}}, \theta)$ denotes rotation by $\theta$ about $\hat{\vc{n}}$.
The corrected robot skeleton $(\shoulder\frm{\robot}, \elbow\frm{\robot}, \wrist\frm{\robot}, \handorientation\frm{\robot})$ is passed to \sewmimic to recover joint angles $\config\frm{\robot}$ in closed form using \sewmimic (\Cref{fig:main_method} (d)); no iterative solver is invoked at any stage.
% \noindent
The 6-DoF torso pose $\pose\frm{\lbl{torso}} = \pose\frm{\human} \cdot \pose\frms{\human}{\robot}$ is solved using a closed-form IK solver \method{ik-geo}~\cite{elias2025ik} in which we always choose the knee-forward solution.

\textbf{Properties.}
\label{sec:ik-multimodality}
\ourmethod provides: \textbf{(i)} a \emph{unique} solution for any non-degenerate human arm pose through the combination of \method{Stereo-sew}, \subproblem{3}, \method{ik-geo}; \textbf{(ii)} \emph{exact} palm-position matching to the human palm by construction (\cref{eq:warp-formulation}); \textbf{(iii)} \emph{whole-body orientation similarity} maximized subject to the EEF constraint via optimal robot placement of Adaptive Offset, the preserved SEW angle $\psi$, and \textbf{(iv)} \emph{closed-form} execution at microsecond speed with no per-robot rescaling or calibration of \sewmimic.
These properties jointly satisfy the precision and consistency requirements of \cref{sec:problem-formulation}.

% \subsection{Task-Priority Switcher}
% \label{sec:task-priority}
% % 4.3 Task-Priority Switcher (~0.5 page)
% %   - Prioritize different body parts to track pose accurately.
% %   \paragraph{VLM-based body-part labeler.} VLM-based detection and labeller.
% %   \paragraph{Smoothing and ramping.} Smoothing and ramping strategy.
% \TODO{for Ye to add}
% Unlike UMI [cite], which retargets only the 6-DoF end-effector pose, our method supports multiple body parts as task-priority targets. We cast task-space selection as classification over a predefined set of body-part classes, with a VLM picking the active target per task. During offline retargeting, we switch goals according to the VLM labels and linearly blend adjacent phases in robot joint space and base SE(2) position. Details in Appendix X. \todo{[TODO]add appendix label}

% \subsection{Lazy Mobile-Base Tracking}
% \label{sec:lower-body}
% 4.4 Lazy Lower-Body Control (~0.5 page)
%   - Stabilize upper-body manipulation while providing mobility support.
%   \paragraph{Formulation.} Idea and formulation.
%   \paragraph{Instantiation on RB-Y1.} Instantiation on RBY1 / RB-Y1.

\textbf{Lazy Mobile-Base Tracking.}\label{sec:lower-body} A whole-body target derived from human data implies a base pose $\mathbf{q}_d\in SE(2)$, but tracking it directly couples every small upper-body shift into wheel motion, and the base's high inertia and limited bandwidth turn those corrections into lag and overshoot that destabilize manipulation. Our key insight is that the $6$-DoF torso can absorb most small upper-body adjustments, so the base should move only for genuine relocation. \ourmethod therefore maintains a \emph{lazy} base target $\mathbf{q}_b$ instead of tracking $\mathbf{q}_d$ directly, and solves torso and arm joints around that target. In effect, the torso handles fine adjustment while the base handles coarse repositioning, improving manipulation stability without sacrificing mobility. Calculation of $\mathbf{q}_b$ are given in \cref{app:lazy-base}.
\begin{wrapfigure}{Hr}{0.2\textwidth}
  \vspace{-1em}
  \centering
  \includegraphics[width=\linewidth]{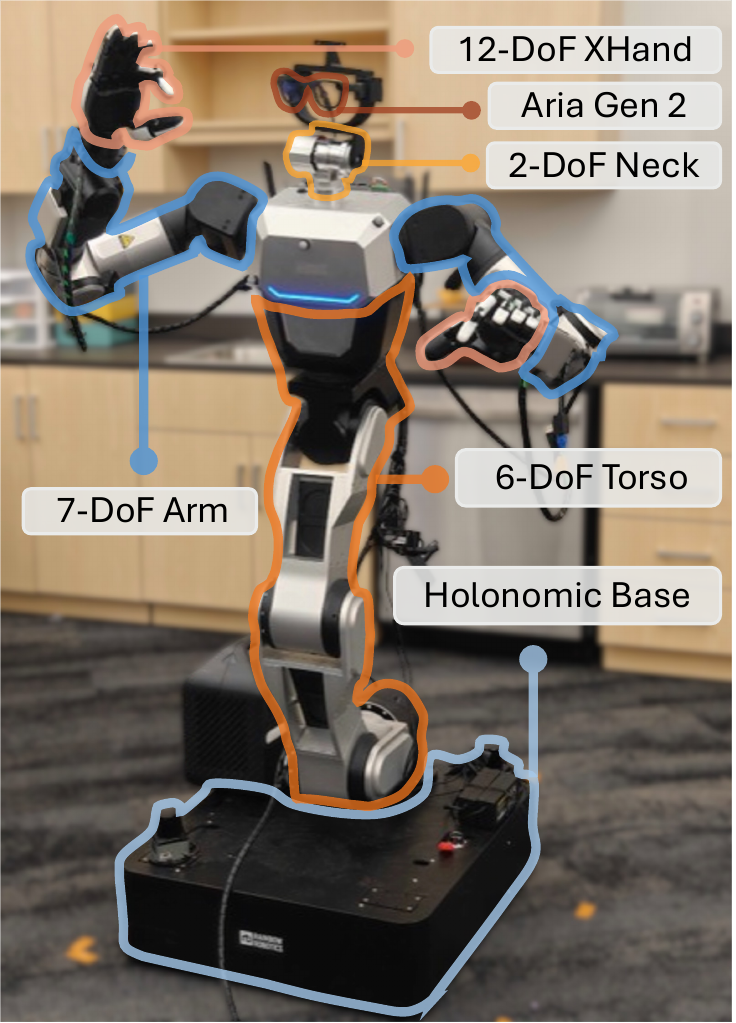}
  \vspace{-1em}
  \caption{\small Our whole-body mobile manipulation platform.}
  \vspace{-1em}
  \label{fig:hardware-intro}
\end{wrapfigure}
% \subsection{Data Collection and Robot System}

\textbf{Data Collection and Robot System}
We collect demonstrations using a single Meta Quest headset, without external motion-capture rigs or robots in the loop. Unlike traditional teleoperation and UMI-style interfaces, which compress a demonstration down to the end-effector’s spatial pose, we capture the operator’s whole-body motion. The system logs the operator's root locomotion alongside a 6-DoF upper-body kinematic tree with full hand articulation in a fixed world frame at 60 Hz (Fig. \ref{fig:teaser}).
% \paragraph{Humanoid Robot platform.}
% \noindent\textbf{Humanoid Robot platform.}
We deploy the RB-Y1 humanoid (Fig.~\ref{fig:hardware-intro}), featuring a holonomic base, 6-DoF torso, two 7-DoF arms, and 12-DoF five-fingered XHands. The robot is commanded via joint-impedance control at 100 Hz. To isolate our kinematic and policy contributions from perception and odometry drift, we track manipulated objects using AprilTags (via a head-mounted Project Aria Gen 2 Glasses~\cite{projectaria2025ariagen2}) and localize the robot base using a Vicon system, providing precise localization required for replay.

\section{Experiments}
\label{sec:experiments}
Our experiments validate that WARP's retargeting solution is precise, consistent, and whole-body human-like, and that these properties make the resulting robot data more learnable--improving whole-body mobile-manipulation policies.
We evaluate the following hypotheses \textbf{H1 Retargeting quality.} \textsc{WARP} produces higher-quality whole-body motion than end-effector-only and generic whole-body retargeting baselines, as measured by tracking accuracy, hardware feasibility, solution consistency, and solving speed on our mobile manipulation platform (\cref{fig:hardware-intro}). \textbf{H2 Downstream learnability.} The motion properties produced by \textsc{WARP} translate into measurably better policy learning. Identical policies trained on \textsc{WARP}-retargeted data outperform those trained on baseline-retargeted data in success rate and rollout consistency. \textbf{H3 Necessity of whole-body retargeting.} For tasks whose success depends on null-space configuration (torso, elbow, or base placement), end-effector-only retargeting is fundamentally insufficient regardless of human data quality and quantity, while \textsc{WARP} succeeds.
%   This establishes that whole-body retargeting is necessary, not merely preferable.

% BONES-SEED provides three motion-format subsets (SOMA Uniform, SOMA Proportional, Unitree G1; see Appendix~\ref{app:seed}). We use SOMA Proportional, which preserves per-actor body proportions and captures realistic human motion geometry. From it, we sample 730 manipulation demonstrations via language search over the metadata.

% \begin{figure}[t] 
%     \centering
%     \includegraphics[width=\textwidth]{figures/radar_test.pdf}
%     \caption{Retargeting feasibility and motion-quality diagnostics for the method. From left to right: hardware feasibility on a linear scale, retargeting and motion quality on a linear scale, hardware feasibility on a log scale, and retargeting and motion quality on a log scale.}
%     \label{fig:method-retargeting-diagnostics}
% \end{figure}

% Required packages:
% \usepackage{booktabs}
% \usepackage[table]{xcolor}
% \usepackage{graphicx}

\subsection{Evaluation on Retargeting Motion Quality}

% \nolinenumbers
\begin{figure}[htbp]
  \centering
  \includegraphics[
    width=1.0\linewidth,
    trim={8pt 5pt 0 10pt},
    clip
  ]{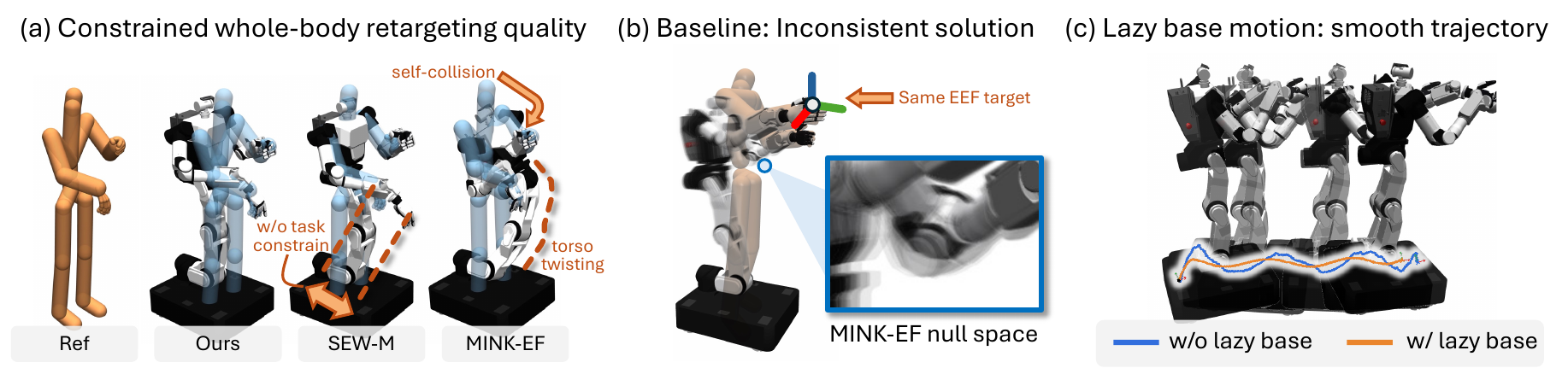}
  \vspace{-15pt}
  \caption{\small \textbf{\ourmethod{} is precise, consistent, and smooth.}
\textbf{(a)} \ourmethod{} matches both the task constraint (end-effector pose) and the human body pose; baselines satisfy only one---SEW-M loses the end-effector, MINK-EF self-collides and twists the torso.
\textbf{(b)} Under a perturbed initial guess with the end-effector target fixed, MINK's optimization is inconsistent, while \ourmethod{}'s closed-form solution is identical every time.
\textbf{(c)} A responsive upper body compensates for the low-frequency \emph{lazy base}, also smoothing the base trajectory.}
  \label{fig:features}
\end{figure}
% \linenumbers

\label{sec:exp_h1}
\begin{figure*}[t]
\centering
\begin{minipage}[c]{0.32\textwidth}
  \centering
  \includegraphics[width=\linewidth]{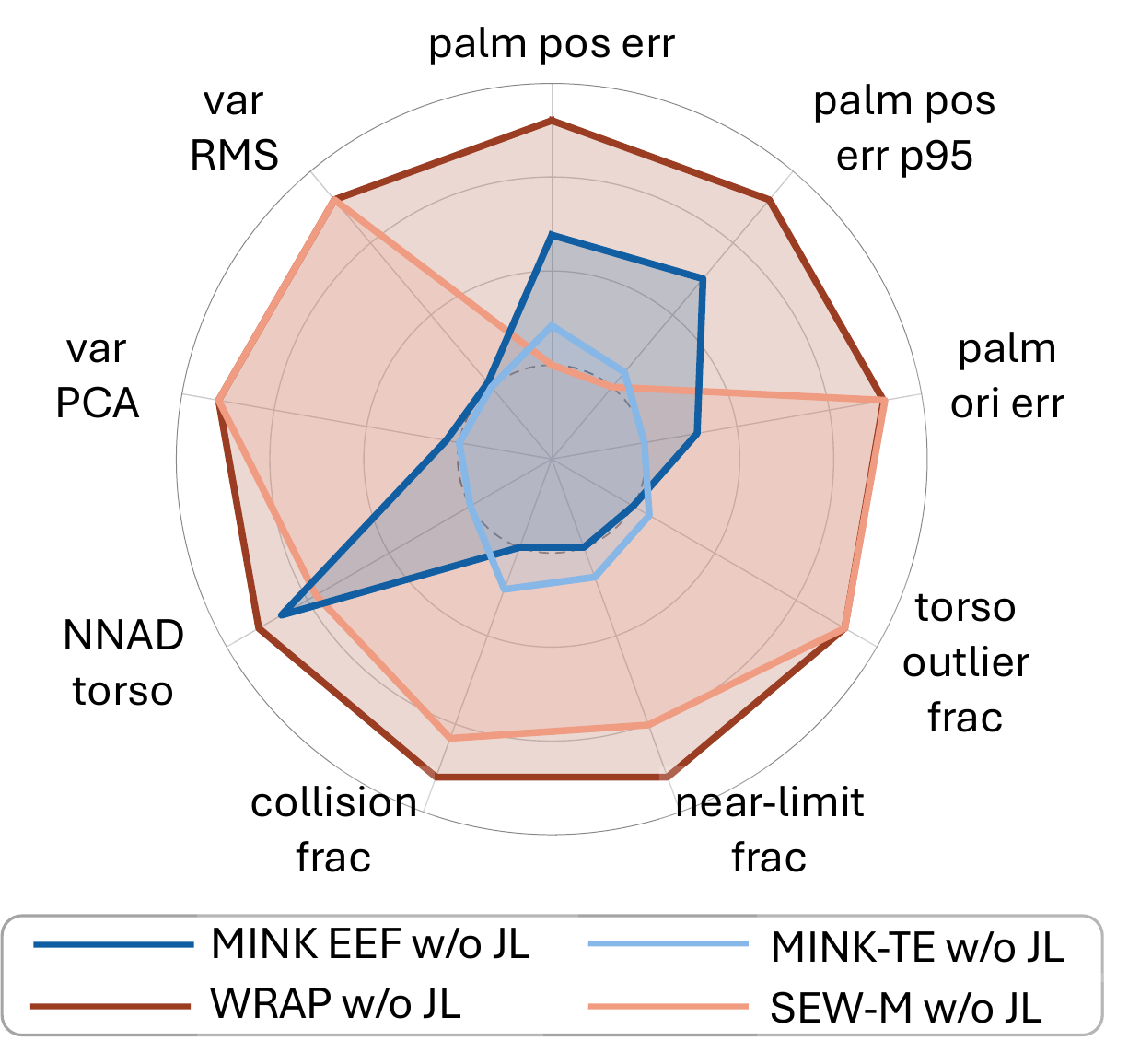}
\end{minipage}
\hfill
\begin{minipage}[c]{0.66\textwidth}
  \centering
  \footnotesize
  \setlength{\tabcolsep}{1.8pt}
  \renewcommand{\arraystretch}{1.06}
  \resizebox{\linewidth}{!}{%
  \begin{tabular}{@{}ll|ccc|ccc|ccc@{}}
  \toprule
  & & \multicolumn{3}{c|}{\textbf{Tracking error} $\downarrow$} & \multicolumn{3}{c|}{\textbf{Feasibility} $\downarrow$} & \multicolumn{3}{c}{\textbf{Solver variation} $\downarrow$} \\
  \cmidrule(lr){3-5} \cmidrule(lr){6-8} \cmidrule(lr){9-11}
  \textbf{Method} & \textbf{JL} & \textbf{Palm} & \textbf{P95} & \textbf{Ori.} & \textbf{Torso} & \textbf{Limit} & \textbf{Coll.} & \textbf{NNAD} & \textbf{PCA} & \textbf{RMS} \\
  & & mm & mm & deg & frac. & frac. & frac. & -- & eig. & deg \\
  \midrule
  SEW-M   & off & 178.979 & 201.056 & \textbf{7.89e-6} & \textbf{0.000} & 0.0126 & 0.243 & 0.488 & 1.20e-25 & 6.83e-14 \\
  MINK-EF & off & \underline{0.701} & \underline{1.853} & 0.0107 & 0.625 & 0.1610 & 0.977 & 0.368 & 173.49 & 3.117 \\
  MINK-TE & off & 18.557 & 73.980 & 0.157 & \underline{0.027} & 0.0852 & 0.640 & 1.026 & 1119.11 & 6.106 \\
  \rowcolor{purple!10}\textbf{WARP} & off & \textbf{0.0046} & \textbf{0.046} & \underline{8.74e-6} & \textbf{0.000} & \textbf{0.0047} & 0.163 & \underline{0.289} & 1.14e-25 & 6.66e-14 \\
  \midrule
  SEW-M   & on  & 215.641 & 272.842 & 6.304 & 0.162 & 0.0131 & \underline{0.084} & 0.442 & \underline{1.09e-25} & \underline{6.56e-14} \\
  MINK-EF & on  & 0.751 & 2.478 & 0.0112 & 0.611 & 0.1247 & 0.222 & 0.354 & 227.29 & 3.169 \\
  MINK-TE & on  & 19.492 & 69.345 & 0.188 & 0.049 & 0.0420 & 0.478 & 0.454 & 613.47 & 4.964 \\
  \rowcolor{purple!10}\textbf{WARP} & on  & 24.048 & 82.036 & 3.259 & 0.130 & \underline{0.0060} & \textbf{0.017} & \textbf{0.266} & \textbf{1.06e-25} & \textbf{6.46e-14} \\
  \bottomrule
  \end{tabular}%
  }
\end{minipage}

\vspace{-0.35em}
\caption{\small Simulation retargeting results. Left: radar visualization of retargeting feasibility and motion-quality diagnostics. Right: quantitative results for the highlighted variants. All metrics are lower-is-better. Best results are shown in \textbf{bold}; second-best results are \underline{underlined}.}
\label{fig:sim-retargeting-results}
\vspace{-2em}
\end{figure*}

% \paragraph{Setup.} 
\textbf{Setup.} 
We evaluate retargeted motion data quality using a high-quality human motion dataset BONES-SEED-SOMA ~\cite{bones_seed_2026}). It preserves per-actor body proportions and captures realistic human motion geometry. From it, we sample 514 manipulation demonstrations via language search over the provided human motion descriptions \cref{app:seed}.
We compare \textsc{WARP} against two retargeting baselines spanning the dominant design points in the literature: \textbf{MINK}~\cite{mink} (optimization-based IK, end-effector-only) MINK-EF, \textbf{body constrained MINK} MINK-TE (end-effector, torso orientation, and elbow swivel angle introduced in Eq.\ref{eq:elbow-robot}), and \textbf{SEW-M}~\cite{kong2026sew} (SEW-Mimic, unconstrained).

\textbf{Metrics.} We evaluate retargeting along three axes, all lower-is-better unless noted. \textbf{Tracking accuracy} captures fidelity to the target human pose via the mean and 95th-percentile palm position error, and the palm orientation error. \textbf{Hardware feasibility} checks whether a kinematical solution is safe on the real robot via the torso outlier rate, the joints near-limit rate, and the self-collision rate. \textbf{Solution consistency}, which matters for policy learning, is measured by NNAD, PCA, and the RMS standard deviation of joint angles. We give all metric formulas in \cref{sec:metrics}.

\textbf{WARP produces accurate, feasible, and consistent retargeting solutions.} It achieves the lowest tracking error among all JL-off methods, as shown in \cref{fig:sim-retargeting-results}: even without auxiliary constraints or regularization, it reduces palm position error by more than \textbf{$150\times$} over MINK-EF and drives hand-orientation error to machine precision, both direct consequences of the closed-form SEW solver, which enforces exact tracking by construction rather than through a weighted cost. (Its JL-on error is larger, but mainly because aggressive motions in the BONE-SEED proportional dataset push some targets outside the robot's reachable workspace under a human-like posture, where less human-like methods like MINK can approach them more closely.) WARP also produces the most feasible motion, with the lowest joint-limit and self-collision fractions: a humanoid's large null space admits many task-satisfying but physically infeasible configurations that exceed joint limits or self-collide, so a retargeter should encode the human pose prior to suppress them, and WARP's human-like solutions do exactly that, acting as a self-collision-avoidance prior that baselines capture less well (Fig.~\ref{fig:features}). It is likewise the most consistent across variants, with the best NNAD, PCA, and RMS solver-variation scores by orders of magnitude; unlike optimization-based approaches, its solutions are deterministic and initial-condition-independent, so the policy learns from consistent, non-conflicting targets rather than noisy, seed-dependent ones. Finally, WARP solves roughly \textbf{$30\times$} faster than iterative optimization-based retargeters---converting SEED takes about an hour on one CPU versus a full day for the baseline---leaving headroom for online use.

\subsection{Simulation Evaluation}

\paragraph{Setup.}
\begin{wrapfigure}{r}{0.4\textwidth}
  \vspace{-12pt}
  \begin{minipage}{0.4\textwidth}
    \centering
    \vspace{-4em}\includegraphics[width=\linewidth]{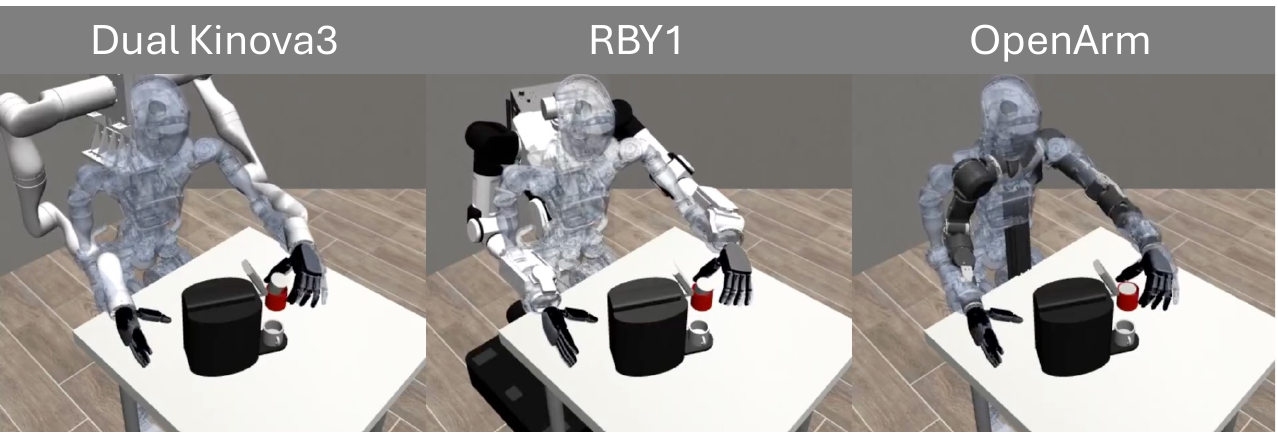}

    \captionof{figure}{\small WARP retargets one robot motion to different robot embodiments.}
    \label{fig:cross_embodiment}
    % \vspace{-10pt}
    {\footnotesize\setlength{\tabcolsep}{1.5pt}%
    \resizebox{\linewidth}{!}{%
    \begin{tabular}{@{}l|cc|cc|cc|cc@{}}
      \toprule
      & \multicolumn{2}{c|}{\textit{can\_sort}} & \multicolumn{2}{c|}{\textit{pouring}} & \multicolumn{2}{c|}{\textit{coffee}} & \multicolumn{2}{c}{\textit{average}} \\
      \cmidrule(lr){2-3} \cmidrule(lr){4-5} \cmidrule(lr){6-7} \cmidrule(lr){8-9}
      Method & replay & policy & replay & policy & replay & policy & replay & policy \\
      \midrule
      MINK & \textbf{99.5\%} & 94\% & 88.5\% & 74\% & 50.5\% & 8\% & 79.5\% & 59\% \\
      \rowcolor{purple!10}
      \textbf{\textit{WARP}} & 98.5\% & \textbf{100\%} & \textbf{90.5\%} & \textbf{78\%} & \textbf{51.0\%} & \textbf{34\%} & \textbf{80.0\%} & \textbf{71\%} \\
      \bottomrule
    \end{tabular}%
    }}
    \captionof{table}{\small Replay and policy rollout result (\%) on DexMimicGen tasks. WARP achieves $12\%$ higher success rate for policy.}
    \label{tab:h1_replay_policy}
  \end{minipage}
  \vspace{-1.5em}
\end{wrapfigure}

DexMimicGen~\cite{jiang2024dexmimicen} synthesizes large dexterous-manipulation datasets from a small set of seed teleoperation demonstrations through task-structured replay with randomized object resets. We use 200 synthesized GR1 demonstrations per task across three bimanual benchmarks---\textit{coffee}, \textit{pouring}, and \textit{can-sort}. For each demonstration, we retarget the motion to the RB-Y1 humanoid with Fourier hands using both WARP and MINK, then filter the results through MuJoCo replay with stochastic retries. Finally, we train one behavior-cloning policy for each task-retargeter pair using the demonstrations that succeed under both retargeters, i.e., the WARP--MINK success intersection.

\textbf{WARP retarget data leads to higher task success.}
At replay of DexMimicGen tasks, WARP and MINK reach comparable success rates---open-loop execution
of either retargeter's output completes the task at similar rates---but the
gap opens during policy training: policies trained on WARP data achieve
substantially higher task-execution success than those trained on MINK data,
with the margin widest on fine-grained manipulation such as the coffee task.
Replay parity therefore understates the difference between retargeters. The
feasibility and solution-consistency properties do not
materially change what is replayable for robot data, but they directly shape what a policy
can learn from the resulting data---so retargeting nuances invisible at replay become decisive downstream.
% We highlight that the data is not wholebody and produces from the teleop, thus the action data is more "robotfridendly"
We further shows that WARP can transfer robot data to different embodiments as shown in \cref{fig:cross_embodiment}.
% \paragraph{Finding3: WARP is robust to multiple embodiment input/output and transfer.}

% \TODO{1. NNAD and position variance (maybe vis) 2. sampling  rate of csv reader will impact the MINK result too we can use this result too.}
\begin{wrapfigure}{r}{0.5\textwidth}
\vspace{-3em}
\centering
\includegraphics[
width=\linewidth,
]{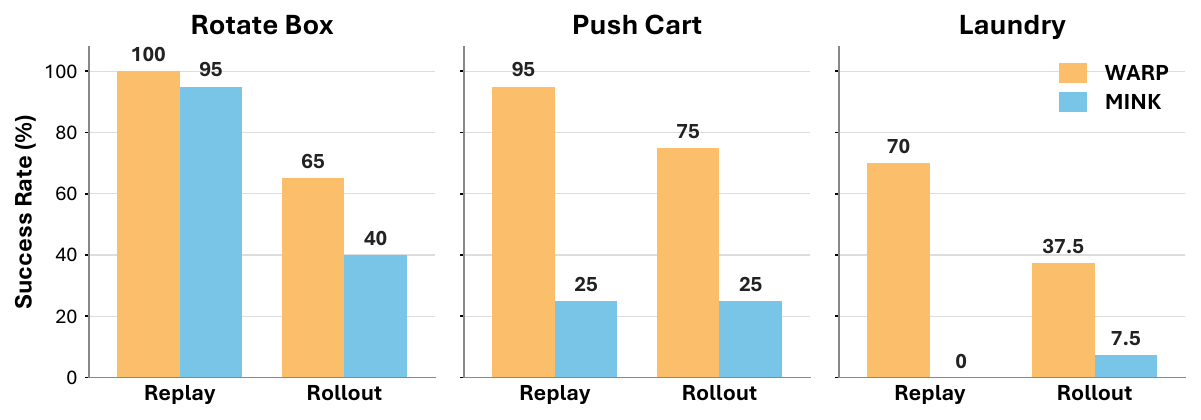}
\vspace{-2em}
\caption{\small Real-world evaluation of retargeted data replay and policy rollout (10 trials each).}
\label{fig:hardware_exp_result}
\vspace{-1.5em}
\end{wrapfigure}

\subsection{Real-World Evaluation}
\label{sec:exp_h2}

\begin{figure}[ht]
    \centering
    \includegraphics[width=\linewidth]{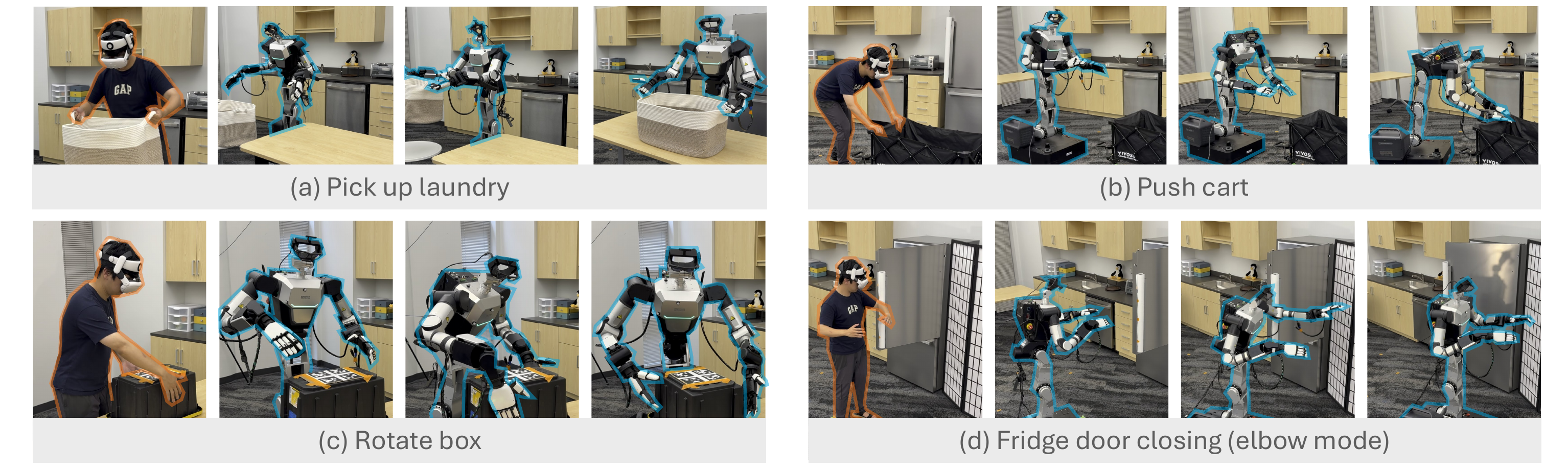}
    \vspace{-1.5em}
    \caption{\small \textbf{\textcolor[HTML]{F17720}{Human demonstrations}} and \textbf{\textcolor[HTML]{00A7E1}{robot executions}} across four real-world tasks.}
    \label{fig:real_world_task_flow}
    \vspace{-1em}
\end{figure}

We collect 50 human demonstrations for each task and train a policy with it \cref{app:policy}. \textbf{Pick-up-laundry.}
The operator lifts a laundry basket by both handles, moves it to an adjacent
table, and sets it down. The task stresses bimanual wrist control: each handle
must be gripped at a precise orientation and that orientation re-achieved to
release, so palm position alone is insufficient.
\textbf{Push-cart.}
The operator pushes a small cart forward by $0.5\,\mathrm{m}$ with both hands.
The task tests base--arm coordination under contact: success depends not just on
where the hands go but on whether the body clears the cart.
\textbf{Rotate-box.}
The operator grips opposite sides of a box and rotates it $90^{\circ}$
clockwise. The task stresses torso--arm coordination, since the robot must twist
its upper body to place both end-effectors at the required orientation.
\textbf{Fridge-door-closing (elbow mode).}
The operator closes a fridge door with both hands occupied, pushing it shut with
the left elbow---contact with a body link other than the hands, outside any
end-effector-only formulation. Without proper observation, we train no policy and evaluate replay, where WARP succeeds in 90\% of trials.

\begin{figure*}[t]
    \centering
    \includegraphics[width=0.95\textwidth]{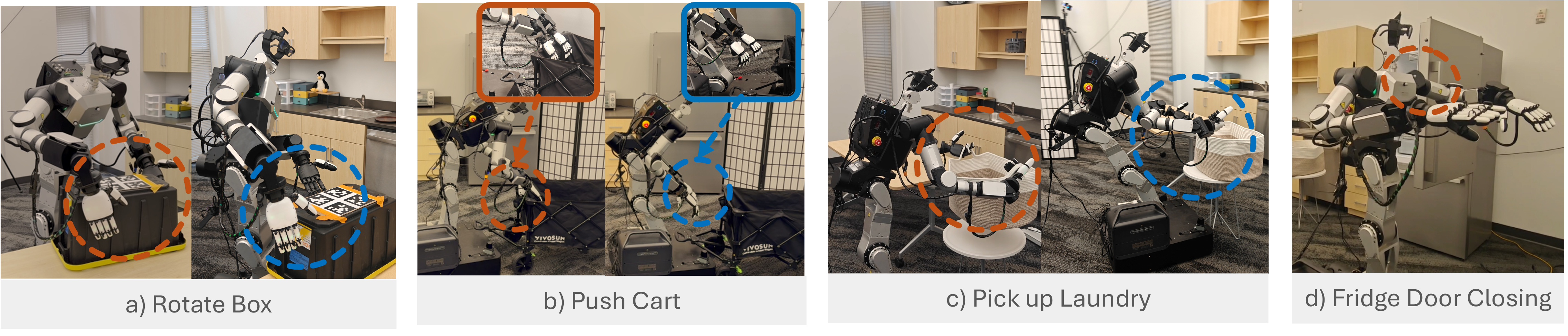}

    \caption{\small {Comparison with \textbf{\textcolor[HTML]{B25628}{WARP}} and \textbf{\textcolor[HTML]{306EBA}{MINK}} on our tasks.  WARP resolves constraints across the whole kinematic chain, preserving palm pose and human-like posture—a motion-quality advantage driving policy success. MINK drives the torso and base into extreme or colliding configurations: a) over-rotating, b) hitting the cart, and c) accumulating wrist-orientation errors that break grasps. d) shows that WARP can also treat the elbow as a task-space objective.}}
    \vspace{-2em}
    \label{fig:exp_case}
\end{figure*}

\textbf{Finding Summary.}
Across all four tasks, WARP outperforms the end-effector-centric MINK baseline
(\cref{fig:hardware_exp_result}). By coordinating torso and arm as the
demonstrator did, WARP keeps palm contact and recoverable posture, whereas MINK
drives the upper body into extreme configurations feasible only at open-loop
replay (rotate-box, pick-up-laundry). Its whole-body objective exploits the full
kinematic chain: WARP supports manipulation with arbitrary links such as the
elbow (fridge-door-closing) and preserves the implicit base--arm constraints
that end-effector- or keypoint-based retargeting discards by construction
(push-cart, verifying~H3). Its closed-form SEW solver (Finding~1) tracks the
hand frame $\mathbf{H}$ to machine precision, guaranteeing the wrist alignment
that grasp and release demand---where MINK's accumulated orientation errors
become decisive (pick-up-laundry). The result is more learnable data: WARP
replays the fridge task in 90\% of trials and, even at comparable replay rates,
trains stronger policies (65\% vs.\ 40\% on rotate-box). Motion quality, not
replay completion, governs downstream success (\cref{fig:exp_case}).

% \subsection{Ablations}
% \label{sec:exp_ablations}

% \paragraph{SEW ablation} We evaluate the two main components, the embodiment offset and the functional offset in the dexmimicgen task / realworld replay task.

% \section{Discussion}
% \label{sec:discussion}
% \input{sections/06_discussion}
\section{Conclusion}
\label{sec:conclusion}

% Conclusion paragraph

We present {WARP}, an offline 
framework of human-to-robot retargeting and policy-learning for whole-body mobile manipulation by leveraging SEW representation between human and robot pose retargeting. 
% it enables a robust system with human motion prior, robust EEF alignment, and more learnable demonstrations.
To our knowledge, WARP is the first system to train a whole-body mobile manipulation policy directly from offline human demonstrations, without teleoperation.This supports further research on policy learning of high-quality human pose data for zero-shot transfer on a tractable subset of real-world tasks.

\textbf{Limitation}
Image observation is lacking for our policy training, limiting the tasks we could try.
We would try image policy in the next step and invesigate how WARP can help in visual-motor policy from offline human demonstration.

\bibliography{references}

\clearpage
\appendix
% \section*{Supplementary Material}

\clearpage
\appendix
\startcontents[appendix]

\section*{Supplementary Material}
\addcontentsline{toc}{section}{Supplementary Material}

% Show down to paragraph level in the local appendix contents
\setcounter{tocdepth}{4}

% \section*{Appendix Contents}
\begingroup
\small

% Optional: make paragraph-level entries compact in the appendix TOC
\titlecontents{paragraph}
  [4.8em]
  {}
  {\contentslabel{3.8em}}
  {}
  {\titlerule*[0.5pc]{.}\contentspage}

\printcontents[appendix]{}{1}{}

\endgroup
\vspace{1em}

% \todo{TODO: Table of contents}

% \todo{TODO: 1. why fail; 1.5 additional results; 2. implementation (1)retargetor, (2)elbow mode, (3)policy details; 3. Experiment details (1 dataset, data coll)} 

% \todo{Dex Mimicgen}

% \todo{Expand Limitations}

% \zc{mention the playback and policy speed is slower than human, 6hz, data 20hz}
% switcher as potential app

% \ye{Method Details & Explanations
% Explain the DexMimicGen experiment setting
% Briefly describe the elbow mode
% Define the policy structure and observation space 
% Add detailed descriptions of the MINK variants
% Explain the MINK null-space figure
% explain why existing methods fail (with qualitative results)
% Qualitative Results
% DexMimicGen qualitative results
% Adaptive offset qualitative results
% Lazy base tracking qualitative results
% Data collection device / UI results
% Additional qualitative examples --- all kinds
% Experimental Results & Metrics
% DexMimicGen replay and policy success rates
% MINK sweep experiments
% Metric definitions
% Radar graph values
% WARP failure cases on BONE-SEED, including why they fail
% Limitations
% Current kinematic limitations of the retargeting method
% Lack of dynamics modeling in simulation
% Cross-References
% Since hyperlinks are unavailable in this version, we should make the appendix searchable by reusing clear keywords from the main text?}

% Use the appendix for extended derivations, implementation details, additional experiments, hardware notes, extra plots, or failure-case analysis.

\section{Why Existing Retargeting Fails?}

Existing whole-body retargeters---e.g., GMR~\citep{araujo2025gmr}, built on
\method{MINK}~\citep{mink}---make the opposite choice on both axes: they optimize
in joint space
% directly in joint-velocity space rather than first constructing a target
% skeleton,
and they track the wrist as a soft objective rather than a hard
constraint. Concretely, they pose retargeting as a weighted multi-task
velocity-level IK,
\begin{equation}
    \dot{\config}^{\ast}
    = \argmin_{\dot{\config}}\;
    \sum_i w_i \norm{J_i\,\dot{\config} - \dot{x}_i^{\ast}}^{2}
    + w_{\mathrm{posture}}\,\norm{\dot{\config} - \dot{\config}_{\mathrm{posture}}}^{2},
    \label{eq:mink}
\end{equation}
where each task term $i$ is a soft whole-body objective---wrist pose, torso
orientation, elbow-swivel angle, or head
orientation introduced in ~\cref{app:mink_variants}---and the final term is posture regularization. Wrist and palm
tracking are thus one weighted term among many, never enforced exactly, and hand
and finger articulation fall outside the formulation entirely.

This conflates cross-embodiment skeleton matching with joint-angle IK, producing
three failure modes that violate the replayability and consistency requirements
above. \textbf{(1)~Inconsistency:} the velocity update descends to whichever
null-space branch is nearest the seed $\config_0$, so different seeds satisfy the
same task yet settle in disjoint regions of joint space, giving a multi-modal
training target that confuses policy learning. \textbf{(2)~Imprecision:} the task
weights couple wrist accuracy to posture---raising the posture weight for
whole-body imitation degrades wrist tracking and breaks replayability, while
lowering it drives the robot toward unnatural or unsafe configurations.
\textbf{(3)~Kinematic Embodiment Gap:} difference between robot-skeleton keypoints and
unscaled human keypoints systematically bias the retargeted poses, and retargeting methods are sensitive to how reference points are mapped~\citep{araujo2025gmr}.

WARP removes (1)~by construction, returning one closed-form solution per pose;
(2)~by promoting palm matching to a hard constraint; and (3)~by operating on
scale-invariant SEW geometry rather than absolute keypoint coordinates.

\subsection{MINK Baseline Variants}
\label{app:mink_variants}

We mainly evaluate two variants of the MINK-based retargeting baseline to separate the effect of end-effector tracking from additional whole-body imitation terms. Both variants use the same MINK solver and formulate retargeting as a weighted velocity-level IK problem over the robot body joints. At each iteration, the solver minimizes the active task-space residuals and integrates the resulting joint velocity to update the robot configuration.  \textbf{MINK's formulation.} Both MINK variants use the same weighted velocity-level IK solver. The end-effector objective tracks the pose of both robot palms, including their 3D positions and orientations. This encourages the robot hands to follow the target palm trajectories extracted from the human motion. The torso objective tracks the orientation of the robot torso against the target upper-body orientation, encouraging whole-body alignment beyond the hands. The elbow objective matches the arm swivel configuration using a scalar SEW angle, rather than directly tracking the 3D elbow position. This provides a compact way to encourage similar arm posture while avoiding over-constraining the elbow location.  The \textbf{MINK-EF} variant only uses the end-effector objective and therefore focuses on accurate palm tracking. The \textbf{MINK-TE} variant additionally enables the torso and elbow objectives, aiming to imitate more of the human whole-body motion. This comparison allows us to isolate whether adding whole-body imitation terms improves retargeting quality beyond palm-pose tracking alone. 

MINK combines these terms as weighted soft objectives,
\begin{equation*}
\dot{\config}^{\ast}
=
\arg\min_{\dot{\config}}
;
w_{\mathrm{EEF}}\mathcal{C}_{\mathrm{EEF}}
+
w_T\mathcal{C}_{T}
+
w_{\psi}\mathcal{C}_{\psi}
+
w_{\mathrm{posture}}\mathcal{C}_{\mathrm{posture}}
\end{equation*}
Finger motion is not optimized by MINK and is instead solved by the analytical XHand IK module.

\paragraph*{MINK-EF.}
MINK-EF is the end-effector-only variant. It activates the left and right palm-pose objectives, with palm position and orientation costs both set to $1.0$. MINK-EF only asks the robot to match the target palm poses, while the remaining redundant degrees of freedom are resolved implicitly by the optimizer, posture regularization, damping, and initialization. This variant tests whether end-effector tracking alone is sufficient for cross-embodiment replay.

\paragraph*{MINK-TE.}
MINK-TE extends MINK-EF by adding soft torso and elbow imitation terms. In addition to the same left and right palm-pose objectives used in MINK-EF, MINK-TE activates a torso orientation objective with cost $0.5$ and a scalar elbow-swivel objective with cost $0.2$. The elbow term does not directly match the human elbow position in 3D. Instead, it matches the stereographic SEW swivel angle $\psi$ computed from the shoulder, elbow, and wrist points.

\begin{table}[t]
\centering
\small
\setlength{\tabcolsep}{15pt}
\renewcommand{\arraystretch}{1.1}
\begin{tabular}{lcc}
\hline
\textbf{Task / parameter} & \textbf{MINK-EF} & \textbf{MINK-TE} \\
\hline
Palm position cost & $1.0$ & $1.0$ \\
Palm orientation cost & $1.0$ & $1.0$ \\
Torso orientation cost & $0.0$ & $0.5$ \\
Elbow-swivel cost & $0.0$ & $0.2$ \\
\hline
\end{tabular}
\vspace{10pt}
\caption{MINK baseline variants used in our paper visualization. MINK-EF tracks only the palm pose, while MINK-TE additionally uses soft torso-orientation and elbow-swivel objectives.}
\label{tab:mink_variants}
\end{table}

For the paper visualization, we use MINK-EF and MINK-TE as the two MINK baselines. MINK-EF represents a minimal end-effector tracking baseline, while MINK-TE represents a stronger MINK baseline with additional torso and elbow posture cues. Both variants remain weighted soft-objective IK methods: palm tracking, torso tracking, and elbow-swivel matching are balanced by task weights rather than enforced as closed-form geometric constraints.

\subsection{MINK Tasks Tradeoff}
\label{app:mink_tradeoff}

% \todo{TODO: change SEW to \ourmethod}
% \ye{AI gen for now, req review}

\begin{figure}[h]
  \centering

  \begin{subfigure}[t]{0.32\textwidth}
    \centering
    \includegraphics[width=\linewidth]{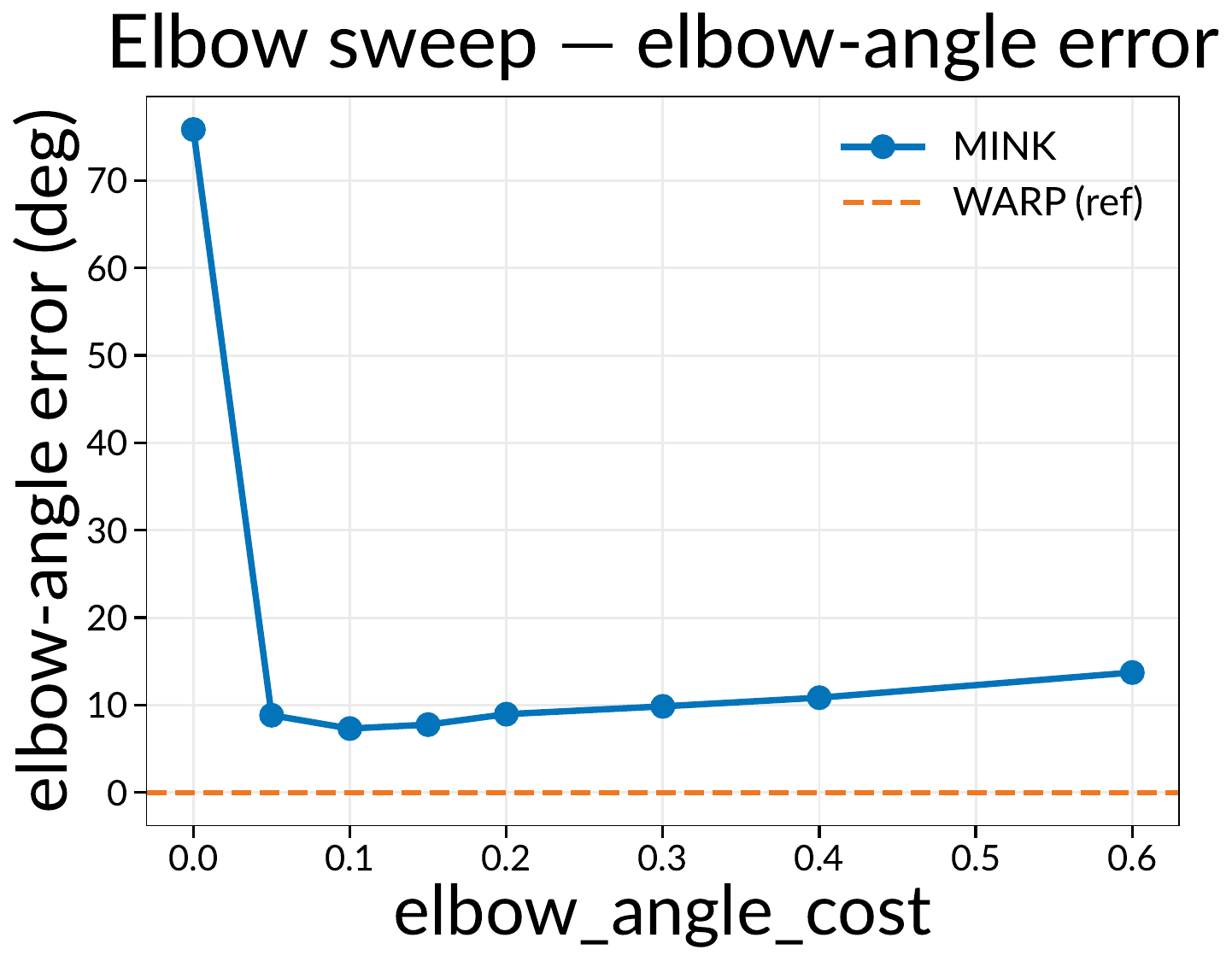}
    \caption{Elbow error}
    \label{fig:sweep-a}
  \end{subfigure}\hfill
  \begin{subfigure}[t]{0.32\textwidth}
    \centering
    \includegraphics[width=\linewidth]{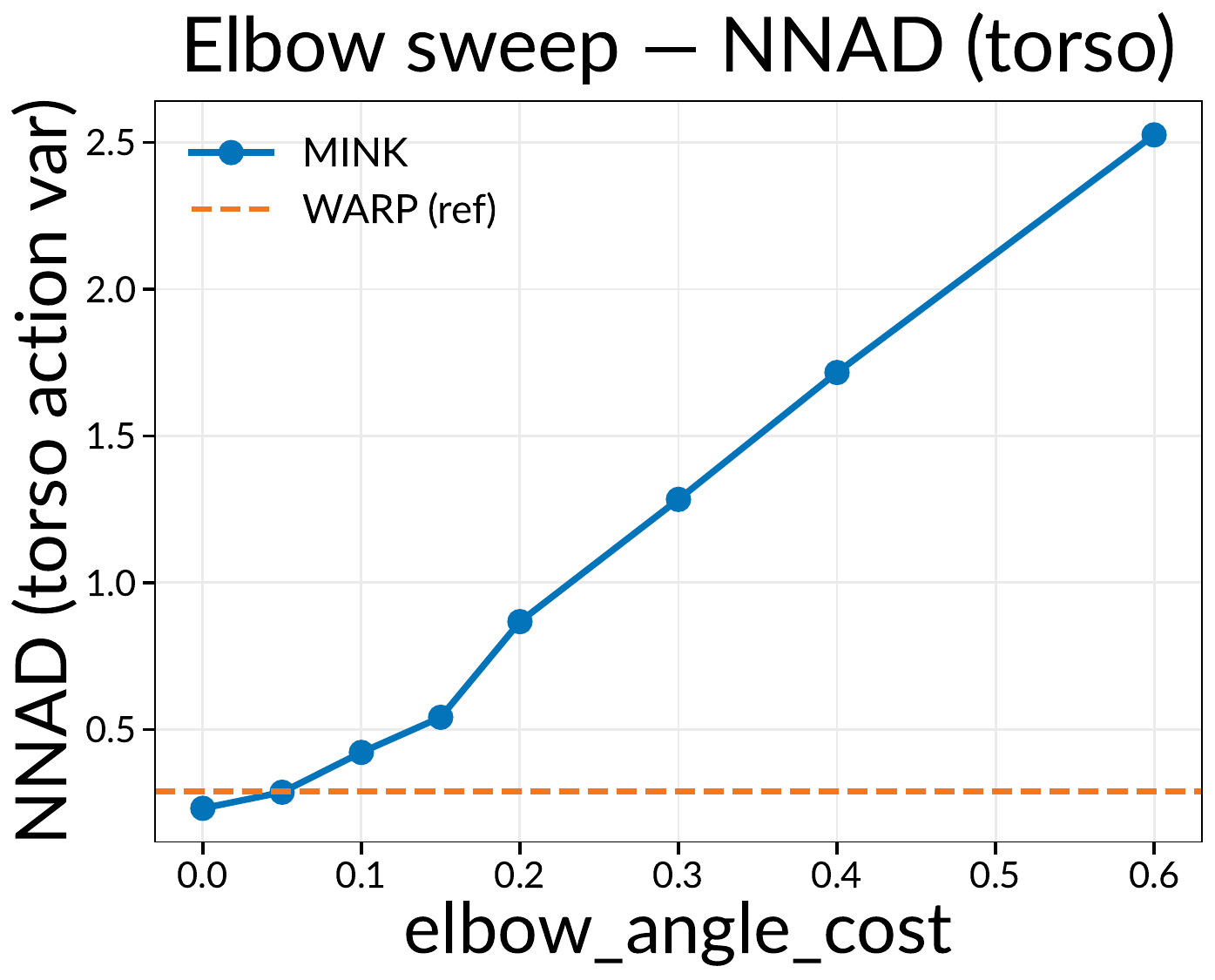}
    \caption{Torso NNAD}
    \label{fig:sweep-b}
  \end{subfigure}\hfill
  \begin{subfigure}[t]{0.32\textwidth}
    \centering
    \includegraphics[width=\linewidth]{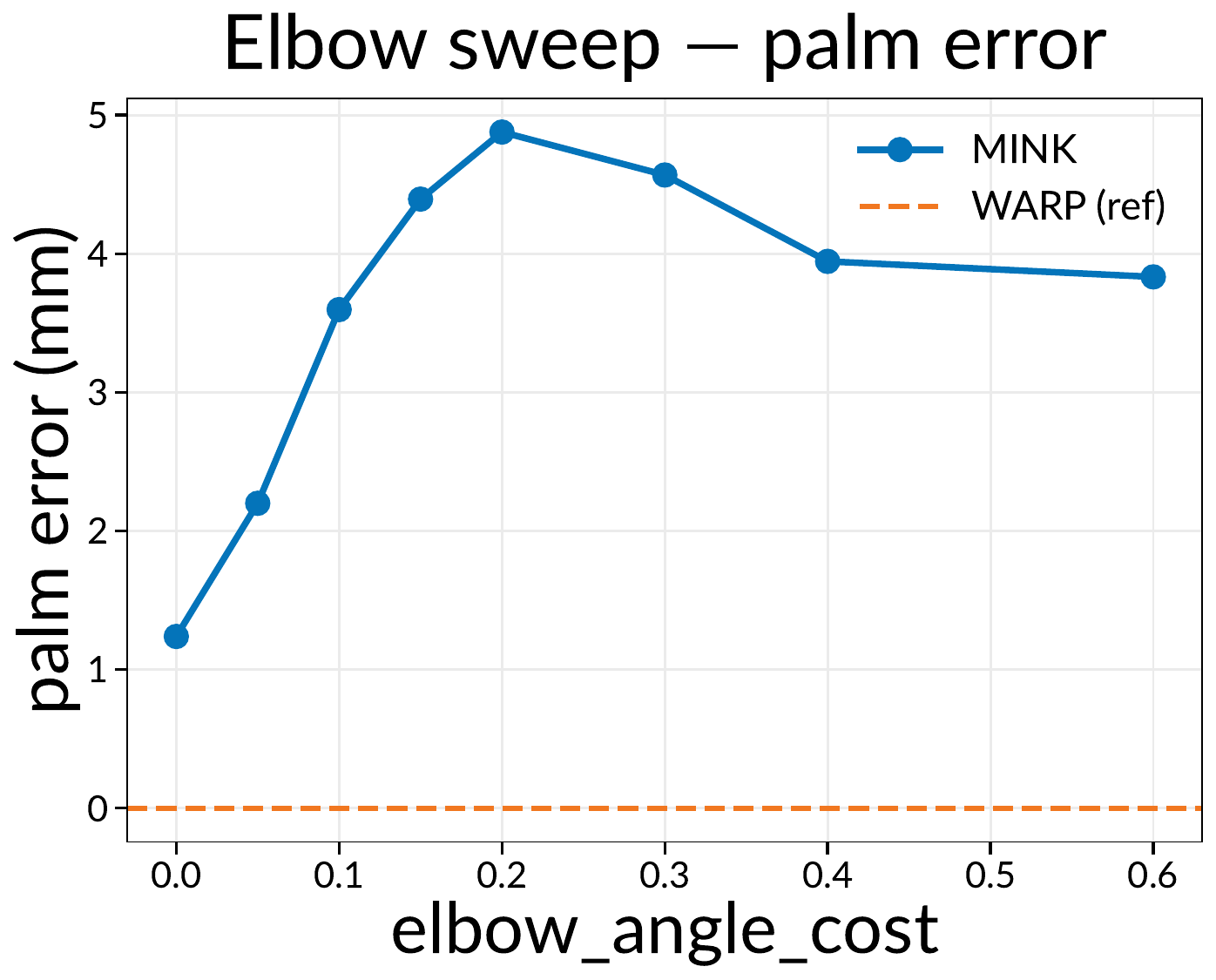}
    \caption{Palm error}
    \label{fig:sweep-c}
  \end{subfigure}

  \vspace{0.5em}

  \begin{subfigure}[t]{0.32\textwidth}
    \centering
    \includegraphics[width=\linewidth]{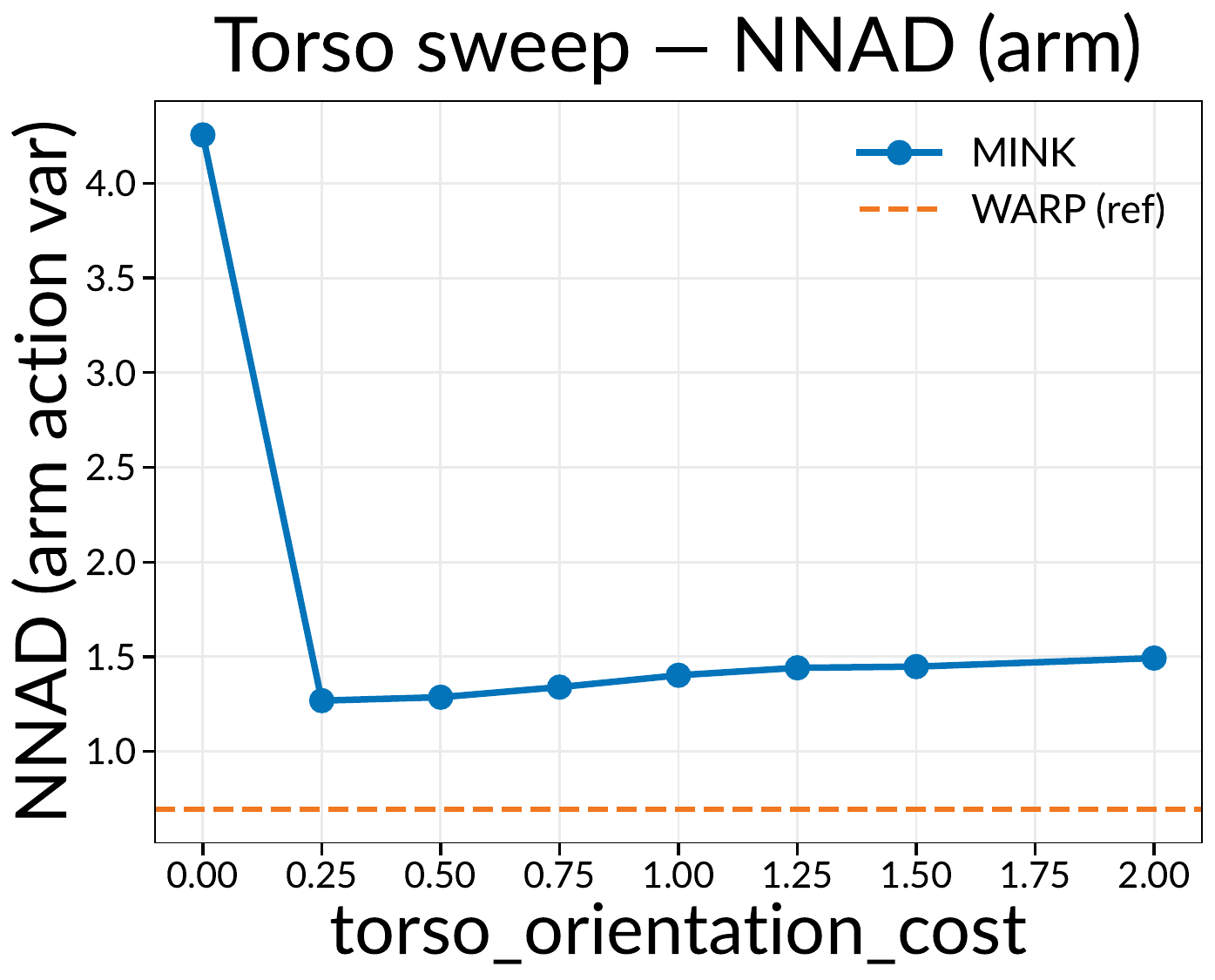}
    \caption{Arm NNAD}
    \label{fig:sweep-d}
  \end{subfigure}\hfill
  \begin{subfigure}[t]{0.32\textwidth}
    \centering
    \includegraphics[width=\linewidth]{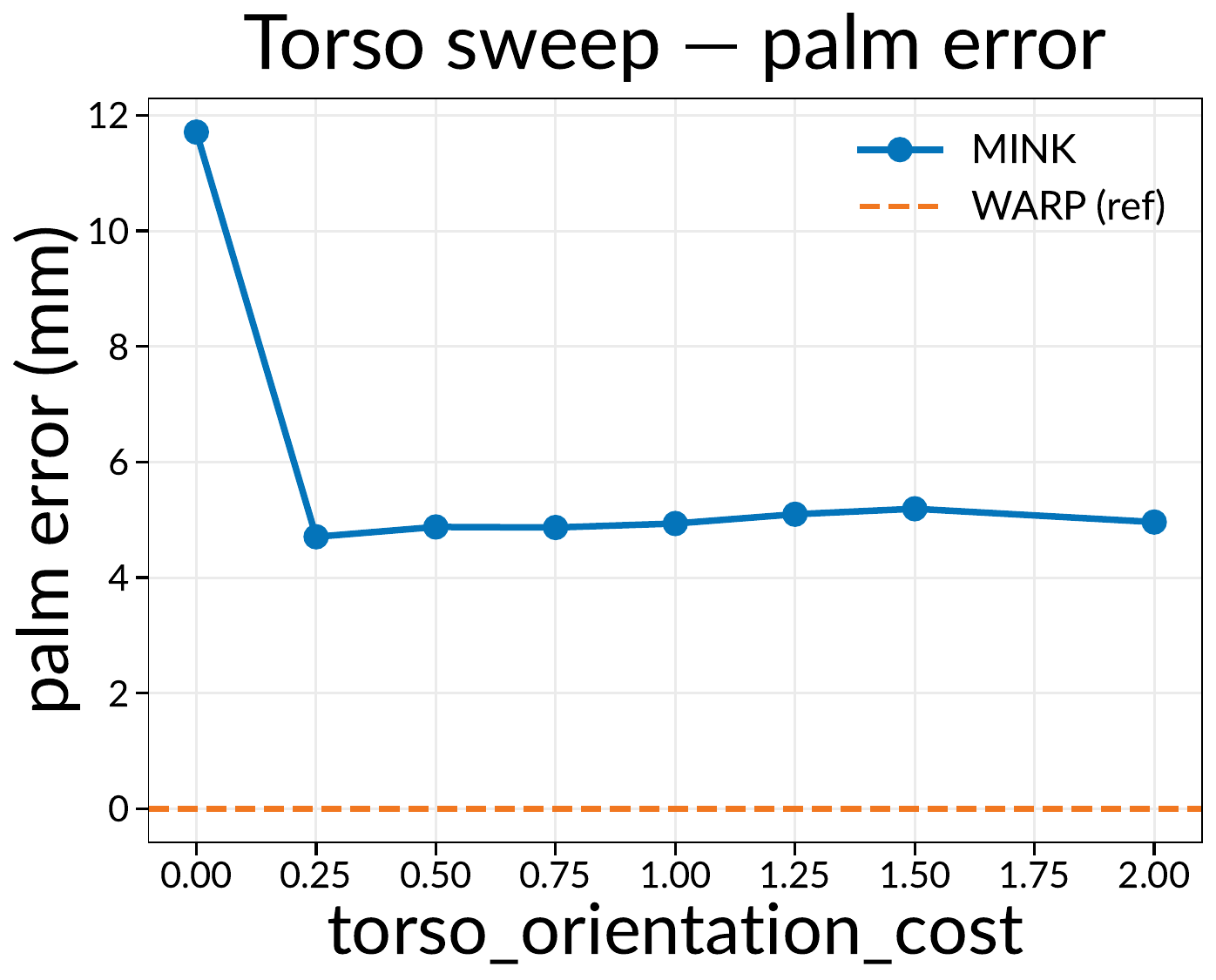}
    \caption{Palm error}
    \label{fig:sweep-e}
  \end{subfigure}\hfill
  \begin{subfigure}[t]{0.32\textwidth}
    \centering
    \includegraphics[width=\linewidth]{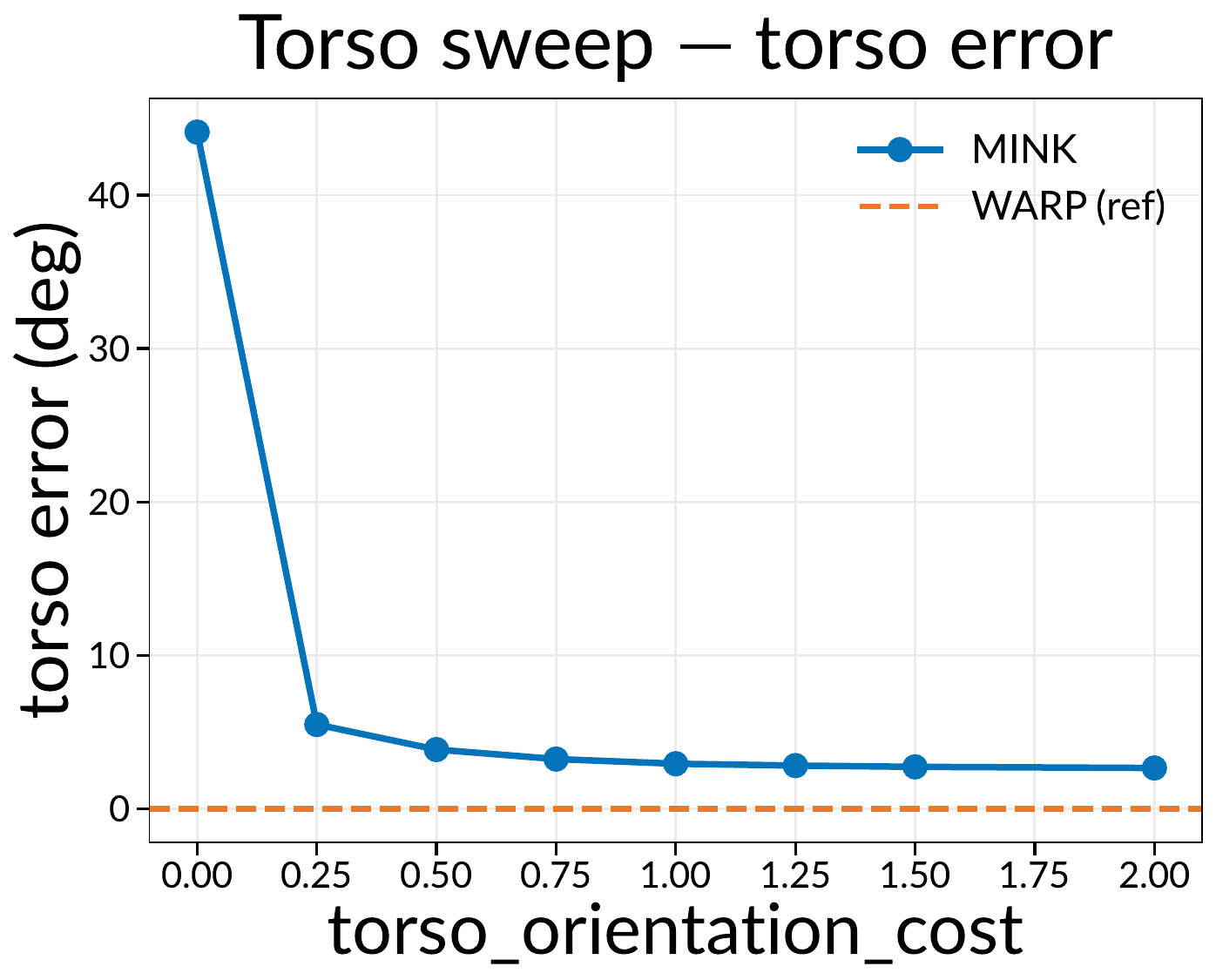}
    \caption{Torso error}
    \label{fig:sweep-f}
  \end{subfigure}

   \caption{\textbf{Posture-cost sweeps: MINK (blue) vs.\ the tuning-free \ourmethod reference
  (orange dashed).} Panels (a)--(c) sweep \texttt{elbow\_angle\_cost}; (d)--(f) sweep
  \texttt{torso\_orientation\_cost}. No single weight matches SEW on palm accuracy, posture
  error, and action consistency at once; see \cref{app:mink_tradeoff}.}
  \label{fig:cost-sweeps}
\end{figure}

MINK exposes posture quality as a soft cost weighted against end-effector (EEF)
tracking, so one scalar weight might trade the two off to some extend; \ourmethod fixes posture analytically
and exposes no such knob. We make this concrete by sweeping each posture weight on the
same demonstrations and reporting palm (EEF) error, the relevant posture error, and NNAD
(action variability) against the tuning-free \ourmethod reference (\cref{fig:cost-sweeps}).

\textbf{Elbow sweep (panels a--c).} At \texttt{elbow\_angle\_cost}$=0$, MINK ignores arm
posture: elbow-angle error reaches $76^\circ$ while palm error stays small ($1.2$\,mm).
Raising the weight drives elbow error to a ${\sim}7^\circ$ minimum near reference, but degrades
both other axes: palm error climbs to a $4.9$\,mm peak at $0.2$, and torso NNAD rises
monotonically from $0.2$ to $2.5$, crossing the WARP level ($0.3$) at ${\sim}0.05$. No weight
recovers \ourmethod's simultaneous near-zero elbow error, palm error, and torso variance.

\textbf{Torso sweep (panels d--f).} Leaving the torso redundancy unregularized
(\texttt{torso\_orientation\_cost}$=0$) is worst on every axis: $44^\circ$ torso error,
$11.7$\,mm palm error, and $4.25$ arm NNAD. A small weight ($0.25$) collapses all three, but
only torso error keeps improving (to ${\sim}3^\circ$); palm error plateaus near $4.7$\,mm and
arm NNAD near $1.3$, both far above the \ourmethod references (${\sim}0$\,mm and $0.7$).

Across both sweeps, every setting that lowers posture error degrades palm accuracy and inflates
NNAD in the other body part, and no setting matches SEW on all three at once. \ourmethod attains the
best value on each metric simultaneously with no per-task tuning: posture quality follows
structurally from the analytic solution rather than from a regularizer. For offline retargeting,
this means MINK forces a per-task choice between accurate and consistent supervision, whereas
\ourmethod supplies both unconditionally.

% \section{Additional Results}

% \todo{add results}

\section{Implementation Details}

\subsection{Body-Centric Frame from Keypoints}
\label{app:make-frame}

We express the SEW skeleton $(\shoulder,\elbow,\wrist,\handorientation,\tool)$ and every
tracked pose in an upper-body-centric frame, so the representation depends only on relative
geometry rather than on where the operator stands or which way they face. The same routine
builds the human frame $\pose\frm{\human}$ and the robot frame $\pose\frm{\robot}$ from each
embodiment's shoulders, giving one shared convention in which the two skeletons are directly
comparable. \cref{alg:make_frame} constructs this frame in closed form at each timestep from
three keypoints.

\paragraph*{Construction.}
The frame is anchored at the shoulder midpoint
$\vc{p}=\tfrac{1}{2}(\shoulder\lside+\shoulder\rside)$, and its orientation is fixed by two
cues. The lateral axis $\vc{u}\lbl{y}$ runs along the shoulder line from the right to the left
keypoint. The remaining axis is pinned by the anchor $\transvec\lbl{anchor}$:
$\vc{p}-\transvec\lbl{anchor}$ gives a coarse trunk reference, and
$\vc{u}\lbl{x}=\normalize\!\bigl(\vc{u}\lbl{y}\times(\vc{p}-\transvec\lbl{anchor})\bigr)$ takes
the direction orthogonal to the shoulder line in the shoulder--anchor plane. With the anchor
below the shoulders, $\vc{u}\lbl{x}$ points anteriorly---the body's facing direction. The
vertical axis $\vc{u}\lbl{z}=\vc{u}\lbl{x}\times\vc{u}\lbl{y}$ completes a right-handed
orthonormal basis $\rotmat=[\vc{u}\lbl{x},\vc{u}\lbl{y},\vc{u}\lbl{z}]$.

The construction is deliberately asymmetric in what it trusts. Only the shoulder line enters
$\rotmat$ exactly; the anchor contributes a single half-plane choice, and its component along
$\vc{u}\lbl{y}$ is removed by the cross product, so anchor noise cannot tilt the frame. The
sole requirement is that the three keypoints be non-collinear---if $\transvec\lbl{anchor}$ fell
on the shoulder line the first cross product would vanish---which holds for any natural
upper-body posture.

\begin{algorithm}[ht]
\caption{Make Frame from Keypoints \\
$\pose \gets \makeframe(\shoulder\lside,\shoulder\rside,\transvec\lbl{anchor})$ \\
\textit{// Given three non-collinear keypoints in $\R^3$, output a coordinate frame pose in $\SE(3)$ parameterized by a rotation matrix and translation vector}}
\label{alg:make_frame}
\begin{algorithmic}[1]
\Require Left, right, and anchor keypoints $\shoulder\lside,\shoulder\rside,\transvec\lbl{anchor}$

\State \textit{// Set translation vector between left / right keypoints}
\State $\vc{p} \gets \tfrac{1}{2}(\shoulder\lside + \shoulder\rside)$

\State \textit{// Represent frame orientation as a rotation matrix}
\State $\vc{u}\lbl{y} \gets \normalize(\shoulder\lside - \shoulder\rside)$
\State $\vc{u}\lbl{x} \gets \normalize(\vc{u}\lbl{y}\times (\vc{p} - \transvec\lbl{anchor}))$
\State $\vc{u}\lbl{z} \gets \vc{u}\lbl{x} \times \vc{u}\lbl{y}$
\State $\rotmat \gets [\vc{u}\lbl{x}, \vc{u}\lbl{y}, \vc{u}\lbl{z}]$

\State $\pose \gets \mat{ \rotmat & \vc{p} \\ \vc{0} & 1 }$

\State \Return homogeneous matrix $\pose$
\end{algorithmic}
\end{algorithm}

\subsection{Elbow Position as Retargeting Goal}

% \ye{ai gen, req proofreading}

WARP first recovers the full robot skeleton
$(\shoulder\frm{\robot},\elbow\frm{\robot},\wrist\frm{\robot},\handorientation\frm{\robot})$
with exact palm matching and the preserved elbow angle $\psi$; this fixes the robot elbow
$\elbow\frm{\robot}$, the intersection of the upper-arm and forearm segments.
We then align the robot elbow to the human elbow on the ground plane only.
Many everyday elbow motions, such as opening or holding a door, depend on the elbow's horizontal placement, not its height, so a 2D projection match captures the relevant constraint without over-constraining the palm-matched arm.

Let $\proj:\mathbb{R}^3\!\to\!\mathbb{R}^2$ project onto the ground plane by dropping the gravity axis.
A planar base shift translates every robot elbow projection equally, so, mirroring the adaptive offset of \cref{eq:adaptive-offset}, we take the shift that aligns the two-arm elbow centroids:
\begin{equation}
\transvec_{\mathrm{elbow}}
=
\tfrac{1}{2}\bigl(
\proj(\elbow\frm{\human}\lside)
+
\proj(\elbow\frm{\human}\rside)
\bigr)
-
\tfrac{1}{2}\bigl(
\proj(\elbow\frm{\robot}\lside)
+
\proj(\elbow\frm{\robot}\rside)
\bigr)
\label{eq:elbow-2d-offset}
\end{equation}
Applying $\transvec_{\mathrm{elbow}}$ sets the desired planar base pose
$\mathbf{q}_d=(\transvec_{\mathrm{elbow}},\theta_d)\in SE(2)$, with yaw $\theta_d$ from the human root heading.
The 6-DoF torso is then re-solved on $\mathbf{q}_d$ by \method{ik-geo} to restore the exact palm constraint of \cref{eq:warp-formulation}: the base provides coarse horizontal placement and the torso absorbs the residual, so the elbow projection is matched without disturbing palm tracking.

\subsection{Potential Elbow-Position Mode Application Extension}
\label{sec:elbow-position-extension}

% \ye{ai gen, req polishment}

WARP's skeleton-first formulation extends beyond palm-centered retargeting to
task-specific priority modes for different body parts. Unlike UMI-style
interfaces, which retarget only the 6-DoF end-effector pose, representing the
full arm skeleton lets the active target be selected from a set of body-part
classes---palm, elbow, torso, or head---which matters when the end-effector
alone does not capture the intended behavior. When pushing against or holding a
door, for example, the horizontal elbow placement, not the palm pose alone,
shapes whether the arm posture is natural and feasible.

As an initial proof of concept, we frame task-target selection as a
classification problem: a VLM labels the task-relevant body part from the task
instruction or demonstration, and the corresponding mode is activated during
offline processing. In elbow-position mode, WARP recovers the full arm skeleton
with exact palm matching and preserved elbow angle as before, then adjusts the
planar base pose to align the robot elbow projection with the human elbow
projection on the ground plane. The hard palm constraint is preserved
throughout, so this base adjustment fixes elbow placement without disturbing
palm tracking.

To avoid discontinuities when the active label changes along a trajectory, we
linearly blend adjacent phases in robot joint space and base $SE(2)$ position,
yielding continuous transitions while preserving the per-phase target.
Elbow-position mode thus illustrates WARP as an application-level extension:
rather than tuning weighted IK costs, we switch the geometric target by task
context. We test only elbow-position mode here as a proof of concept; a
systematic evaluation of VLM-driven mode selection across tasks and body parts
remains future work.

\subsection{Subproblems used in WARP}
% \todo{TODO: Review added subproblems}
\label{app:subproblems}

% \ye{ai gen, req review}

WARP's geometric solver is built on the geometric
\emph{subproblems}~\citep{elias2025ik}: canonical primitives that map a small set
of vectors, axes, and scalars to the joint angle(s) satisfying a fixed geometric
relation, each solved in closed form with \textsc{atan2} and elementary
operations so the result is exact and singularity-robust. The cross-embodiment
elbow placement of \cref{eq:elbow-robot} uses one of them, \subproblem{3}.

\paragraph*{Subproblem 3 (circle and sphere).}
Given vectors $p_1, p_2 \in \R^3$, a unit axis $k \in \R^3$, and a nonnegative
scalar $d$, \subproblem{3} returns the angle $\theta$ that minimizes
$\bigl|\,\norm{R(k,\theta)\,p_1 - p_2} - d\,\bigr|$. Geometrically,
$R(k,\theta)\,p_1$ sweeps a circle about $k$, and the solutions are its
intersections with the sphere of radius $d$ centered at $p_2$: at most two exact
angles, with a continuous least-squares angle when the circle and sphere do not
meet.

\paragraph*{Use in WARP.}
After per-arm palm alignment fixes the robot shoulder $\shoulder\frm{\robot}$ and
wrist $\wrist\frm{\robot}$ and transfers the elbow half-plane normal
$\hat{\vc{n}}$ from the human, the only remaining unknown is the elbow
$\elbow\frm{\robot}$, which must lie at upper-arm length $\length_{SE}$ from the
shoulder, at forearm length $\length_{EW}$ from the wrist, and in the half-plane
of $\hat{\vc{n}}$. Rotating the upper-arm vector $\length_{SE}\,\hat{\vc{e}}_{SW}$
about $\hat{\vc{n}}$, where
$\hat{\vc{e}}_{SW} = \normalize(\wrist\frm{\robot} - \shoulder\frm{\robot})$,
sweeps the circle of candidate elbows at radius $\length_{SE}$ about the
shoulder; the valid elbow is where this circle meets the sphere of radius
$\length_{EW}$ about the wrist. This is exactly \subproblem{3} with
\[
  p_1 = \length_{SE}\,\hat{\vc{e}}_{SW}, \quad
  k   = \hat{\vc{n}}, \quad
  p_2 = \wrist\frm{\robot} - \shoulder\frm{\robot}, \quad
  d   = \length_{EW},
\]
for which $R(k,\theta)\,p_1 - p_2 = \elbow\frm{\robot} - \wrist\frm{\robot}$, so
the \subproblem{3} objective $\norm{R(k,\theta)\,p_1 - p_2} = d$ enforces
$\norm{\elbow\frm{\robot} - \wrist\frm{\robot}} = \length_{EW}$ exactly. Of the at
most two roots we keep $\theta_{SEW} > 0$, the branch matching the human elbow
half-plane, and recover $\elbow\frm{\robot}$ via \cref{eq:elbow-robot}.

The downstream solvers WARP invokes are themselves subproblem-based---
\sewmimic~\citep{kong2026sew} recovers the arm joints from the corrected skeleton
via \subproblem{1} and \subproblem{2}, and \method{ik-geo}~\citep{elias2025ik}
solves the torso pose---but the cross-embodiment alignment introduces only the
single \subproblem{3} call above.

\subsection{Lazy Mobile-Base Filter}
\label{app:lazy-base}

The goal of lazy base tracking is to separate \emph{what} the base should do from \emph{how} small upper-body motion is absorbed. In the main text, the feature is that the torso handles fine adjustment while the base moves only for genuine relocation; here we specify the filter that implements that behavior. Let $\mathbf{q}_d = (\mathbf{p}_d, \theta_d) \in SE(2)$ be the base pose implied by the human upper-body target and $\mathbf{q}_b = (\mathbf{p}_b, \theta_b)$ the filtered base target maintained by \ourmethod{}. With deadband radii $\delta_{xy}, \delta_\theta$, define the deadbanded error
\begin{align}
\tilde{\mathbf{e}}_{xy} &= \max\!\big(0,\, \norm{\mathbf{p}_d - \mathbf{p}_b} - \delta_{xy}\big)\,\hat{\mathbf{e}}_{xy}, &
\tilde{e}_\theta &= \operatorname{sign}(e_\theta)\,\max\!\big(0,\, |e_\theta| - \delta_\theta\big),
\label{eq:lazy-deadband}
\end{align}
where $\hat{\mathbf{e}}_{xy}$ is the unit error direction and $e_\theta$ the wrapped yaw error. The filter then integrates
\begin{align}
\ddot{\mathbf{p}}_b &= \omega_n^2\,\tilde{\mathbf{e}}_{xy} - 2\zeta\omega_n\,\dot{\mathbf{p}}_b, &
\ddot{\theta}_b &= \omega_n^2\,\tilde{e}_\theta - 2\zeta\omega_n\,\dot{\theta}_b,
\label{eq:lazy-filter}
\end{align}
with natural angular frequency $\omega_n = 2\pi f_n$ and damping ratio $\zeta$. Inside the deadband the spring force vanishes and the base coasts to rest under damping alone, so jitter and small sway never reach the wheels; outside it, the spring smoothly draws the base toward the target. We use $\zeta = 1$ for non-overshooting tracking, $f_n \approx 1.5$\,Hz, and deadband radii $\delta_{xy} = 0.05$\,m and $\delta_\theta = 0.1$\,rad.

\subsection{Real Robot System}

% \subsubsection{Hardware specs} 

We deploy the RB-Y1 humanoid (\cref{fig:hardware-intro}), whose kinematics mirror the human upper body:
a holonomic mobile base, a 6-DoF torso, and two 7-DoF arms. The torso lets the upper body reposition without moving the base, improving manipulation stability and extending the reachable workspace beyond an arm-only configuration. Each arm terminates in a 12-DoF five-fingered XHand, of which an internal pose retargeter was used to convert hand poses to robot finger motions. We command the robot through joint-impedance control at $100\,\mathrm{Hz}$ over a wired Ethernet connection.
Visual observations come from a head-mounted Meta Aria Gen~2 sensor streaming at $10\,\mathrm{Hz}$. Because this work does not address the visual embodiment gap, we track each manipulated object with an AprilTag instead of raw pixels, using the Quest cameras during data collection and the Aria cameras during rollout, and express its pose in the robot upper-body frame (\Cref{app:make-frame}). Open-loop replay has no corrective feedback and is therefore sensitive to state-estimation error, so to decouple execution from wheel-odometry drift we measure robot pose with a Vicon motion-capture system. This provides the precise localization that replayability (\cref{sec:problem-formulation}) requires.

\subsection{Policy}
\label{app:policy}

We instantiate our policy using HPT, a transformer-based action-chunk model that
represents observations and actions as tokens and predicts the flow-matching velocity field
over future actions. We use HPT mainly as a standard policy backbone, and introduce a
body-hierarchy bias that mirrors the structure of our retargeting pipeline.

Specifically, we decompose each action chunk into base, torso, arm, and hand components,
\[
\mathbf{a}_{1:H}
=
[\mathbf{a}^{b}, \mathbf{a}^{\tau}, \mathbf{a}^{r}, \mathbf{a}^{h}]_{1:H}.
\]
Rather than treating all body blocks as fully exchangeable, we impose a proximal-to-distal
ordering
\[
b \preceq \tau \preceq r \preceq h,
\]
where proximal motion provides coarse support and reachability, while distal motion refines
interaction. The noisy action chunk is tokenized along both time and body block, and we
apply a block-causal attention mask
\[
M_{ij} =
\begin{cases}
0, & \text{if } j \preceq i, \\
-\infty, & \text{otherwise}.
\end{cases}
\]
No causal mask is applied along the temporal horizon, allowing the model to reason over the
full action chunk jointly. This preserves the standard HPT flow-matching update and
inference cost, while encouraging proximal-to-distal information flow in whole-body action
prediction.
In implementation, low-dimensional observations and hand states are first encoded with
MLP encoders and then fused with the action tokens. We train the policy with a batch size
of 32 and evaluate the checkpoint from epoch 599. The final model remains lightweight,
with approximately 11M trainable parameters.

% =====================================================================
%  Retargeting evaluation metrics — formulas and definitions
%  Auto-generated from projects/xhand_teleop/metrics/kernels/*.py
%  Requires: amsmath, amssymb, bm (optional)
% =====================================================================

\subsection{Evaluation Metrics}
\label{sec:metrics}
\paragraph*{Metrics Definition}
\label{app:metric-definitions}
% Add this near the beginning of the appendix section, or in the preamble.
\newcommand{\metricdef}[2]{\noindent\textit{#1}. #2\par\vspace{0.25em}}

We report 18 metrics covering palm tracking, body-pose tracking, joint-limit feasibility, collision, motion deviation, solver variation, and solve time. All position errors are reported in millimetres, angular errors in degrees, and solve time in milliseconds per frame. Unless stated otherwise, each metric is first computed per demonstration over valid post-warmup frames, then averaged across demonstrations using a NaN-safe macro mean, so each demonstration has equal weight regardless of length.

\medskip
\noindent\textbf{Notation.}
Let $\mathcal{V}$ be the valid frames of a demonstration and $\langle\cdot\rangle$ the NaN-safe macro mean across demonstrations. For arm $s\in\{\mathrm{L},\mathrm{R}\}$, $e_t=\tfrac12\sum_{s}\lVert \bm{p}^{r}_{s,t}-\bm{p}^{h}_{s,t}\rVert$ is the per-frame palm error; $Q_p$ denotes the $p$th percentile; $\angle(R_a,R_b)=\arccos\!\big(\tfrac{1}{2}(\operatorname{tr}(R_a^{\top}R_b)-1)\big)$ the geodesic angle between two rotations; $\mathrm{wrap}(\cdot)$ maps an angle to $[-\pi,\pi]$; and $\bm{q}\in\mathbb{R}^{26}$ the robot joint vector with body-group slices $\bm{q}^{g}$.

\medskip
\noindent\textbf{Palm tracking.}
For each frame and each arm, we compute the Euclidean distance between the robot palm and the corresponding human palm in the demo-world frame, $e_t$.

\metricdef{Raw palm error}{Mean palm position error over all valid post-warmup frames, $\big\langle \operatorname{mean}_{t\in\mathcal{V}} e_t\big\rangle$. This measures average palm tracking accuracy but includes catastrophic outlier frames.}

\metricdef{Clean palm error}{Mean palm error after removing frames whose error exceeds $\tau$, $\big\langle \operatorname{mean}_{t\in\mathcal{V},\,e_t\le\tau} e_t\big\rangle$. This reflects typical-frame tracking quality; a large Raw--Clean gap indicates that a small number of failed frames dominate the raw average.}

\metricdef{P95 palm error}{The $95$th percentile of palm error over valid post-warmup frames, $\big\langle Q_{95}(\{e_t\}_{t\in\mathcal{V}})\big\rangle$. This captures the upper tail of the palm tracking error.}

\metricdef{P99 palm error}{The $99$th percentile, $\big\langle Q_{99}(\{e_t\}_{t\in\mathcal{V}})\big\rangle$. This highlights rare but severe tracking failures.}

\medskip
\noindent\textbf{Arm and torso tracking.}
These metrics evaluate whether the retargeted motion preserves the human elbow and torso configuration, rather than only matching the palm pose.

\metricdef{Elbow error}{Mean absolute error between the robot and human SEW elbow swivel angles $\psi$, $\big\langle \operatorname{mean}_{t\in\mathcal{V}} |\mathrm{wrap}(\psi^{r}_t-\psi^{h}_t)|\big\rangle$. Lower values indicate better preservation of the human elbow configuration around the shoulder--wrist axis.}

\metricdef{Torso error}{Mean rotation-angle error between the robot torso and the human upper-body orientation, $e^{\mathrm{to}}_t=\angle(R^{r}_t,R^{h}_t)$, giving $\big\langle \operatorname{mean}_{t\in\mathcal{V}} e^{\mathrm{to}}_t\big\rangle$. Lower values indicate closer torso pose matching.}

\metricdef{Out@15}{Fraction of valid post-warmup frames whose torso error exceeds $15^\circ$, $\big\langle \operatorname{mean}_{t\in\mathcal{V}} \mathbb{1}[e^{\mathrm{to}}_t>15^\circ]\big\rangle$. This measures how often the retargeted motion enters a visibly poor torso-pose state.}

\medskip
\noindent\textbf{Joint-limit feasibility.}
For each joint $j$ with mechanical limits $[\ell_j,u_j]$, the per-frame margin is $m_{j,t}=\min(q_{j,t}-\ell_j,\,u_j-q_{j,t})$. Wheels and joints without known mechanical bounds are excluded.

\metricdef{Limit fraction}{Fraction of joint--frame pairs within $10^\circ$ of a mechanical limit, $\big\langle \operatorname{mean}_{j,t}\mathbb{1}[m_{j,t}<10^\circ]\big\rangle$. Lower values indicate fewer near-limit configurations.}

\metricdef{Minimum margin}{Minimum margin over all joints and valid post-warmup frames, macro-averaged across demonstrations, $\big\langle \min_{j,\,t\in\mathcal{V}} m_{j,t}\big\rangle$. Larger is better; a negative value means the planned trajectory exceeds the robot's mechanical limits and would be clipped on hardware.}

\medskip
\noindent\textbf{Self-collision.}
This metric evaluates physical feasibility by replaying the retargeted robot trajectory in simulation.

\metricdef{Collision fraction}{Fraction of valid post-warmup frames in self-collision, $\big\langle \operatorname{mean}_{t\in\mathcal{V}} c_t\big\rangle$ with $c_t\in\{0,1\}$. We check the arms, torso, head, and wheels, while keeping the fingers in a neutral pose. Lower values indicate safer trajectories for hardware deployment.}

\medskip
\noindent\textbf{Normalized motion deviation.}
We use Nearest-Neighbor Action Disagreement (NNAD) to measure trajectory-stack determinism. For each human-state observation $i$, let $\mathcal{N}_k(i)$ be its $k$ nearest human-state neighbours drawn from other demonstrations; the group-$g$ disagreement is
\begin{equation}
  \mathrm{NNAD}_g \;=\; \Big\langle \operatorname*{mean}_{i}\ \tfrac{1}{k}\!\!\sum_{j\in\mathcal{N}_k(i)}\!\!\big\lVert \bm{q}^{g}_i-\bm{q}^{g}_j\big\rVert \Big\rangle .
\end{equation}
Low NNAD means that similar human inputs map to similar robot outputs, which is important for stable downstream policy learning.

\metricdef{R-arm NNAD}{$\mathrm{NNAD}_g$ for the right-arm joints. Lower values indicate more deterministic right-arm retargeting.}

\metricdef{L-arm NNAD}{$\mathrm{NNAD}_g$ for the left-arm joints. Lower values indicate more deterministic left-arm retargeting.}

\metricdef{Torso NNAD}{$\mathrm{NNAD}_g$ for the torso joints. Lower values indicate more consistent torso retargeting.}

\metricdef{Head NNAD}{$\mathrm{NNAD}_g$ for the head joints. Lower values indicate more consistent head motion.}

\medskip
\noindent\textbf{Solver variation.}
These metrics come from a controlled pose-perturbation study. For each sampled input pose, the IK solver is run $N$ times with Gaussian-jittered warm starts, producing outputs $\{\bm{q}^{(n)}\}_{n=1}^{N}$; $\langle\cdot\rangle_{p}$ denotes the mean over sampled poses. Unlike the previous metrics, these isolate solver-only non-determinism.

\metricdef{RMS variation}{Root-mean-square per-joint standard deviation across repeated trials (degrees), $\big\langle \sqrt{\tfrac{1}{J}\sum_{j}\operatorname{Var}_n q^{(n)}_j}\big\rangle_{p}$. This measures typical per-joint solver uncertainty.}

\metricdef{Pair-L2 variation}{Average pairwise $\ell_2$ distance between repeated outputs, $\big\langle \operatorname{mean}_{a<b}\lVert\bm{q}^{(a)}-\bm{q}^{(b)}\rVert\big\rangle_{p}$. This measures the typical disagreement between two solver outputs for the same input pose.}

\metricdef{PCA variation}{Largest eigenvalue of the output covariance $\Sigma=\operatorname{Cov}_n(\bm{q}^{(n)})$, $\big\langle \lambda_{\max}(\Sigma)\big\rangle_{p}$. This measures how strongly solver variation concentrates along a dominant null-space direction.}

\medskip
\noindent\textbf{Solve time.}
This metric measures solver-only computational cost.

\metricdef{Solve time}{Mean time spent inside the solver per frame, excluding file I/O, simulation bookkeeping, and video writing. Lower values indicate faster retargeting.}

All metrics are better when lower, except \textit{Minimum margin}, where higher is better.

\section{Experiment Details}

\subsection{Experiment Setup}

\subsubsection{Real World Environment Setup}
All data is collected with 20hz from Quest
During the rollout, to accommodate the controller limited performance, we lower the rollout/control freq to 6hz (around 1/3) of the original speed. This setup is applied to both MINK and \ourmethod.

\paragraph{Human data collection}

We collect demonstrations with a single Meta Quest headset, with no external
motion-capture rig, body-worn markers, or robot in the loop. Unlike traditional
teleoperation and UMI-style interfaces, which compress a demonstration to the
end-effector's spatial pose, we capture the operator's whole-body motion: the
headset's inside-out tracking returns, per frame, an upper-body kinematic tree
together with both hands, each bone a 6-DoF pose in a fixed world frame, plus
the root pose, which captures in-place locomotion. The same headset tracks an
AprilTag on the manipulated object in its camera frame, and the simultaneous
skeleton estimate lets us re-express the object in the operator's upper-body
frame. The operator views the scene through passthrough and toggles recording
from the in-headset interface (Fig.~\ref{fig:teaser}); we log this full-body
pose, hand articulation included, at 60~Hz. The resulting raw human dataset is
\[
\dataset\lbl{raw}
=
\set{\left(\pose\frms{\human}{\object}\foft,\, \humandata\foft\right)}_{t=1}^{N},
\]
where $\pose\frms{\human}{\object}\foft \in \SE(3)$ is the object pose in the
operator's upper-body frame and $\humandata\foft \in \R^{d_\humandata}$ is the
human motion feature (hand and finger poses).

Expressing the object in the upper-body frame is what lets each human
observation transfer to the robot without a world frame. Because our
cross-embodiment alignment places a robot at the anchor with its end-effector
coinciding with the operator's, the fixed anchor transform yields the object
pose in the robot's equivalent upper-body frame, $\pose\frms{\robot}{\object}
\foft$ --- exactly what the robot would observe while reproducing the motion. We
retarget each frame to the robot embodiment with \ourmethod, mapping the human
motion feature to a matched robot proprioceptive state $\tilde s_t \in \R^{d_s}$
and a control action $u_t \in \R^{d_a}$. Combined with the frame-transformed
object pose $\pose\frms{h_c}{o}\foft \in \SE(3)$ from
\Cref{sec:problem-formulation}, this yields the robot training dataset
\[
\dataset\lbl{robot}
=
\set{\left(\left(\pose\frms{h_c}{o}\foft,\, \tilde s_t\right),\, u_t\right)}_{t=1}^{N}.
\]

At deployment, the robot computes $\pose\frms{h_c}{o}\foft$ from its own onboard
perception and reads the proprioceptive feature $s_t$ from its own state, then
queries the policy for the next action
\[
a_t = \pi_\theta(o_t),
\qquad
o_t = \left(\pose\frms{\human}{\object}\foft,\, s_t\right).
\]
Because $s_t$ and $\tilde s_t$ share the same space, and
$\pose\frms{\human}{\object}\foft$ is computed identically at training and
deployment, the policy trained on human data transfers to the robot without
fine-tuning.

% \paragraph{Replay Experiment Setup}

\paragraph{Replay Experiment Setup}
We retarget offline human motion into 20\,Hz robot trajectories.
Both open-loop replay and policy rollout run at a 6\,Hz command rate---30\% of
the recorded speed---which keeps the joint-impedance controller's steady-state
tracking error negligible; replay executes the full 20\,Hz trajectory at this
rate, and policies are trained on the same 20\,Hz data.

% \todo{TODO: }
% \ye{mannually reset sounds suspicious here, help find a way to state this}
A motion-capture system provides accurate, drift-free localization of the robot
base. Open-loop replay carries no perception feedback to correct an initial
mismatch, so each trial must begin from the same robot--object relative
configuration the retargeting assumes, including the adaptive offset WARP applies
across embodiments. We establish this configuration once and reproduce it across
trials by fixing the robot start pose and object placement in the mocap frame
with ground markers.

% 1. we record original human data roughly at ~60hz( we set max frequence of quest recording as 60 hz) and when doing retargeting, we downsample with original speed at 20 hz from the human bones data and do offline retargeting to make it robot data at 20 hz. when doing replay, we run all the data points in the 20hz robot data in a lower frequence (reduce to 6hz for replay, equals to slow it down to 30\% to make sure the controller error is ingoreable. ) For policy traing, we use all the steps in the 20hz data, but when inference we also execute in 6 hz, (slow down the robot, also reudce the steady state error produced by controller acceptable.)

% 2. we use a mocap system to provide accurate and unbaised robot base odometry. And we mannually reset robot-to object relative position considering the functional offset we applied when retargeting and later fixed robot starting points and object placement in mocap frame by marking the position on the ground.

\subsubsection{Real-world Per-Task Scoring Criteria}
\label{app:scoring}

Each manipulation task is scored on a graded success scale; the replay-only
fridge task is scored as a binary completion rate.

\textbf{Pick-up-laundry.} We score grasp and release symmetrically: $1.0$ when
both hands complete the phase, $0.5$ for one, $0$ otherwise.

\textbf{Push-cart.} We score on a graded scale: $1.0$ for a clean two-handed
push, $0.5$ for single-handed, $0.25$ for a push with incidental body contact,
and $0$ for missing or hitting the cart without moving it.

\textbf{Rotate-box.} Progress is discretized by rotation angle:
\(60^\circ\!-\!90^\circ = 1.0\), \(30^\circ\!-\!60^\circ = 0.5\),
\(0^\circ\!-\!30^\circ = 0\).

% \subsection{BONE-SEED Dataset Sample}
% \label{app:seed}

% \ye{SEED detailed intro, should be in appendix} The BONES-SEED dataset provides three motion-format subsets. SOMA Uniform standardizes all motions to a single skeleton for consistent topology and scale, supporting large-scale processing and retrieval. SOMA Proportional preserves per-actor body proportions via shape parameters, capturing morphology-dependent motion geometry. Unitree G1 retargets trajectories to the G1 humanoid as MuJoCo-compatible joint sequences for simulation-based imitation learning. We use SOMA Proportional: its actor-specific morphology better preserves real human motion geometry than the uniform representation, making it the right fit for studying human-to-robot retargeting.

\subsection{DexMimicgen Experiment setup}
\label{sec:dexmimicgen_setup}

% \ye{ai gen, req double check}

We evaluate \ourmethod and MINK retargeting on the same GR1 DexMimicGen demonstrations and convert them into RBY1 policy-training data. The source demonstrations are collected from three bimanual tasks: \texttt{two\_arm\_coffee}, \texttt{two\_arm\_pouring}, and \texttt{two\_arm\_can\_sort\_random}. Each demonstration contains the recorded GR1 MuJoCo state, action commands, and environment metadata. We first retarget each GR1 demonstration into a body-centric representation containing shoulder, elbow, wrist, torso, head, and hand information. For SEW, the extracted shoulder--elbow--wrist geometry is later solved online by the RBY1 SEW solver during playback. For MINK, the corresponding RBY1 joint targets are precomputed and stored, so data collection does not require online IK solving.

During collection, each retargeted trajectory is replayed in a dual-robot robosuite environment. The RBY1 robot executes the retargeted motion, while the original GR1 motion is optionally overlaid as a ghost robot for visualization. The robot base is fixed using a canonical base placement computed from a reference frame, which avoids per-demonstration base drift and keeps the comparison between \ourmethod and MINK consistent. Finger retargeting is bypassed: the original 6-DoF Fourier hand command from GR1 is copied directly and deterministically expanded by the Fourier hand controller into the actuator command used by RBY1.

We keep only successful demonstrations for policy training. A trajectory is considered successful only if the task succeeds during execution and remains successful at the final frame. After collection, we further intersect the \ourmethod and MINK datasets using the original demonstration index, so both methods are trained and evaluated on matched demonstrations. Finally, we slice a balanced subset of the first $N=50$ demonstrations for each task and method to construct the final policy-training datasets.

\subsubsection{BONES-SEED Dataset Retargeting Setup}
\label{app:seed}

% \subsubsubsection{BONES-SEED Dataset}

% \ye{SEED detailed intro, should be in appendix}
BONES-SEED~\cite{bones_seed_2026} is a large-scale skeletal human motion dataset for
everyday humanoid behaviors, released in three motion-format subsets. \emph{SOMA Uniform}
standardizes all motions to a single skeleton for consistent topology and scale, supporting
large-scale processing and retrieval. \emph{SOMA Proportional} preserves per-actor body
proportions through shape parameters, capturing morphology-dependent motion geometry.
\emph{Unitree G1} retargets trajectories to the G1 humanoid as MuJoCo-compatible joint
sequences for simulation-based imitation learning. We use SOMA Proportional: its
actor-specific morphology preserves real human motion geometry more faithfully than the
uniform representation, the right fit for studying human-to-robot retargeting.

\paragraph{Dataset sampling.}
We evaluate on a curated subset of the SOMA Proportional motions, drawn from the full SEED
metadata by three filters. A \emph{taxonomy} filter selects motions by the activity labels provided in the original
BONES-SEED metadata: we keep the \texttt{Everyday} package (5{,}816 motions), which the dataset
subdivides into the five categories \texttt{Household}, \texttt{Unusual Locomotion},
\texttt{Consuming}, \texttt{Object Interaction}, and \texttt{Environments}. A \emph{description}
filter then restricts to manipulation-relevant motions: we concatenate each motion's four
natural-language description fields (the \texttt{content\_natural\_desc} columns; the short and
technical descriptions are not searched), lowercase the text, and keep the motion if it contains
any of the substrings \texttt{table}, \texttt{everyday}, or \texttt{object} and none of
\texttt{sit} or \texttt{seated}. Matching is case-insensitive substring containment rather than
whole-word, so \texttt{table} also matches \textit{tabletop} and \textit{vegetables} while
\texttt{sit} also removes \textit{transit} or \textit{position}; this loose rule is intentional
and mostly captures additional table-adjacent motions. Finally, a \emph{filename} filter drops
high-reaching motions---those whose filename contains \texttt{\_high\_}, such as reaching into
high cupboards or fridges---since these often exceed RB-Y1's vertical workspace and would measure
reach-envelope violations rather than retargeting quality. We apply no mirror filtering: original
and mirrored sequences both pass whenever they satisfy the predicates, preserving paired left-
and right-handed variants. The final subset contains 514 motions---257 original and 257
mirrored---comprising 135,896 retargeted robot frames at the IK solver output rate.

\subsection{Additional Results}

\subsubsection{BONE-SEED subset Results across all metrics}
% \ye{put complete results here}
\begin{table*}[htbp]
\centering
\small
\setlength{\tabcolsep}{2.4pt}
\renewcommand{\arraystretch}{1.05}

% =========================
% Row 1: tracking + feasibility
% =========================
\resizebox{0.95\textwidth}{!}{%
\begin{tabular}{@{}l c|cccc|ccc|cc@{}}
\toprule
& & \multicolumn{4}{c|}{\textbf{Palm tracking} $\downarrow$}
& \multicolumn{3}{c|}{\textbf{Arm / torso tracking} $\downarrow$}
& \multicolumn{2}{c}{\textbf{Joint-limit / feasibility}} \\
\cmidrule(lr){3-6}
\cmidrule(lr){7-9}
\cmidrule(lr){10-11}
\textbf{Method} & \textbf{JL}
& \shortstack{\textbf{Raw}\\mm}
& \shortstack{\textbf{Clean}\\mm}
& \shortstack{\textbf{P95}\\mm}
& \shortstack{\textbf{P99}\\mm}
& \shortstack{\textbf{Elbow}\\deg}
& \shortstack{\textbf{Torso}\\deg}
& \shortstack{\textbf{Out@15}\\frac.}
& \shortstack{\textbf{Limit}\\frac.}
& \shortstack{\textbf{Margin} $\uparrow$\\deg} \\
\midrule
SEW-M & off
& 178.979 & -- & 201.056 & 203.800
& \underline{2.669} & \textbf{1.189e-6} & \textbf{0.000}
& 0.0126 & 7.108 \\

MINK-EF & off
& \underline{0.701} & 0.609 & \underline{1.853} & \underline{3.389}
& 79.110 & 20.680 & 0.6247
& 0.1610 & -54.95 \\

MINK-TE & off
& 18.557 & 1.360 & 73.980 & 115.400
& 8.966 & 3.857 & \underline{0.0266}
& 0.0852 & -70.11 \\

\rowcolor{purple!10}\textbf{WARP} & off
& \textbf{0.0046} & \textbf{0.0005} & \textbf{0.0465} & \textbf{0.0621}
& \textbf{0.0105} & \underline{1.220e-6} & \textbf{0.000}
& \textbf{0.0047} & \textbf{13.02} \\
\midrule
SEW-M & on
& 215.641 & -- & 272.842 & 285.700
& 5.775 & 5.329 & 0.1624
& 0.0131 & 4.804 \\

MINK-EF & on
& 0.751 & 0.520 & 2.478 & 4.298
& 39.250 & 19.430 & 0.6113
& 0.1247 & -1.000 \\

MINK-TE & on
& 19.492 & 1.352 & 69.345 & 105.300
& 6.881 & 5.128 & 0.0491
& 0.0420 & 1.648 \\

\rowcolor{purple!10}\textbf{WARP} & on
& 24.048 & \underline{0.154} & 82.036 & 100.600
& 3.344 & 4.451 & 0.1298
& \underline{0.0060} & \underline{9.559} \\
\bottomrule
\end{tabular}%
}

\vspace{0.65em}

% =========================
% Row 2: collision + NNAD + solver variation + solve time
% =========================
\resizebox{0.95\textwidth}{!}{%
\begin{tabular}{@{}l c|c|ccc|ccc|c@{}}
\toprule
& & \multicolumn{1}{c|}{\textbf{Collision} $\downarrow$}
& \multicolumn{3}{c|}{\textbf{Normalized motion deviation} $\downarrow$}
& \multicolumn{3}{c|}{\textbf{Solver variation} $\downarrow$}
& \multicolumn{1}{c}{\textbf{Solve time} $\downarrow$} \\
\cmidrule(lr){3-3}
\cmidrule(lr){4-6}
\cmidrule(lr){7-9}
\cmidrule(lr){10-10}
\textbf{Method} & \textbf{JL}
& \shortstack{\textbf{Coll.}\\frac.}
& \shortstack{\textbf{R-arm}\\--}
& \shortstack{\textbf{L-arm}\\--}
& \shortstack{\textbf{Torso}\\--}
& \shortstack{\textbf{RMS}\\deg}
& \shortstack{\textbf{Pair-L2}\\--}
& \shortstack{\textbf{PCA}\\eig.}
& \shortstack{\textbf{Solve}\\ms} \\
\midrule
SEW-M & off
& 0.243
& \underline{0.6578} & \underline{0.6713} & 0.4879
& 6.828e-14 & \textbf{0} & 1.198e-25
& 14.83 \\

MINK-EF & off
& 0.977
& 0.6951 & 0.7052 & 0.3677
& 3.117 & 19.43 & 173.5
& 192.1 \\

MINK-TE & off
& 0.640
& 1.603 & 1.776 & 1.026
& 6.106 & 34.33 & 1119.0
& 235.8 \\

\rowcolor{purple!10}\textbf{WARP} & off
& 0.163
& 0.6825 & 0.7103 & \underline{0.2888}
& 6.657e-14 & \textbf{0} & 1.140e-25
& 6.781 \\
\midrule
SEW-M & on
& \underline{0.084}
& \textbf{0.6527} & \textbf{0.6567} & 0.4419
& \underline{6.561e-14} & \textbf{0} & \underline{1.089e-25}
& \textbf{4.978} \\

MINK-EF & on
& 0.222
& 0.7233 & 0.7321 & 0.3543
& 3.169 & 19.61 & 227.3
& 192.5 \\

MINK-TE & on
& 0.478
& 0.9570 & 0.9569 & 0.4539
& 4.964 & 28.38 & 613.5
& 234.4 \\

\rowcolor{purple!10}\textbf{WARP} & on
& \textbf{0.017}
& 0.7270 & 0.7434 & \textbf{0.2659}
& \textbf{6.456e-14} & \textbf{0} & \textbf{1.062e-25}
& \underline{5.430} \\
\bottomrule
\end{tabular}%
}

\vspace{-0.4em}
\caption{\small Quantitative simulation retargeting results for the highlighted variants. 
The table is split into two rows for readability: the top row reports tracking and joint-limit feasibility metrics, while the bottom row reports collision, normalized motion deviation, solver variation, and solve time. 
For all metrics, lower is better except \textbf{Margin}, where larger is better. 
Best results are shown in \textbf{bold}; second-best results are \underline{underlined}.}
\label{tab:sim-retargeting-full-metrics}
\vspace{-1.5em}
\end{table*}

\cref{tab:sim-retargeting-full-metrics} shows complete metrics results on the 514 trajectories subset.

\subsubsection{Random Sample BONE-SEED Dataset Results}

% \ye{to add the big table}

\begin{table*}[htbp]
\centering
\small
\setlength{\tabcolsep}{2.4pt}
\renewcommand{\arraystretch}{1.05}

% =========================
% Row 1: tracking + feasibility
% =========================
\resizebox{0.95\textwidth}{!}{%
\begin{tabular}{@{}l c|cccc|ccc|cc@{}}
\toprule
& & \multicolumn{4}{c|}{\textbf{Palm tracking} $\downarrow$}
& \multicolumn{3}{c|}{\textbf{Arm / torso tracking} $\downarrow$}
& \multicolumn{2}{c}{\textbf{Joint-limit / feasibility}} \\
\cmidrule(lr){3-6}
\cmidrule(lr){7-9}
\cmidrule(lr){10-11}
\textbf{Method} & \textbf{JL}
& \shortstack{\textbf{Raw}\\mm}
& \shortstack{\textbf{Clean}\\mm}
& \shortstack{\textbf{P95}\\mm}
& \shortstack{\textbf{P99}\\mm}
& \shortstack{\textbf{Elbow}\\deg}
& \shortstack{\textbf{Torso}\\deg}
& \shortstack{\textbf{Out@15}\\frac.}
& \shortstack{\textbf{Limit}\\frac.}
& \shortstack{\textbf{Margin} $\uparrow$\\deg} \\
\midrule
SEW-M & off
& \underline{200.2} & -- & \underline{252.0} & \underline{317.8}
& \underline{2.672} & \underline{1.060} & \underline{0.01311}
& \underline{0.02643} & \textbf{-4.891} \\

MINK-EF & off
& 248.0 & \textbf{0.7856} & 432.0 & 450.4
& 83.43 & 39.46 & 0.7340
& 0.1603 & -61.74 \\

MINK-TE & off
& 310.3 & 1.586 & 654.4 & 797.3
& 33.25 & 17.85 & 0.2221
& 0.1890 & -270.0 \\

\rowcolor{purple!10}\textbf{WARP} & off
& \textbf{11.91} & \underline{1.51} & \textbf{43.77} & \textbf{78.98}
& \textbf{1.144} & \textbf{0.6469} & \textbf{0.00706}
& \textbf{0.02343} & \underline{-7.736} \\
\bottomrule
\end{tabular}%
}

\vspace{0.65em}

% =========================
% Row 2: collision + NNAD + solver variation + solve time
% =========================
\resizebox{0.95\textwidth}{!}{%
\begin{tabular}{@{}l c|c|ccc|ccc|c@{}}
\toprule
& & \multicolumn{1}{c|}{\textbf{Collision} $\downarrow$}
& \multicolumn{3}{c|}{\textbf{Normalized motion deviation} $\downarrow$}
& \multicolumn{3}{c|}{\textbf{Solver variation} $\downarrow$}
& \multicolumn{1}{c}{\textbf{Solve time} $\downarrow$} \\
\cmidrule(lr){3-3}
\cmidrule(lr){4-6}
\cmidrule(lr){7-9}
\cmidrule(lr){10-10}
\textbf{Method} & \textbf{JL}
& \shortstack{\textbf{Coll.}\\frac.}
& \shortstack{\textbf{R-arm}\\--}
& \shortstack{\textbf{L-arm}\\--}
& \shortstack{\textbf{Torso}\\--}
& \shortstack{\textbf{RMS}\\deg}
& \shortstack{\textbf{Pair-L2}\\--}
& \shortstack{\textbf{PCA}\\eig.}
& \shortstack{\textbf{Solve}\\ms} \\
\midrule
SEW-M & off
& \textbf{0.1028}
& \textbf{1.061} & \textbf{1.014} & 1.125
& \underline{6.828e-14} & \textbf{0} & \underline{1.198e-25}
& \textbf{7.626} \\

MINK-EF & off
& 0.9203
& \underline{1.266} & 1.266 & \textbf{0.7478}
& 3.117 & 19.43 & 173.5
& 382.7 \\

MINK-TE & off
& 0.4639
& 5.089 & 5.128 & 3.370
& 6.106 & 34.33 & 1119.0
& 452.1 \\

\rowcolor{purple!10}\textbf{WARP} & off
& \underline{0.1112}
& 1.298 & \underline{1.234} & \underline{0.7921}
& \textbf{6.657e-14} & \textbf{0} & \textbf{1.140e-25}
& \underline{10.52} \\
\bottomrule
\end{tabular}%
}

\vspace{-0.4em}
\caption{\small Quantitative retargeting results on the 7,623-motion BONE-SEED two-per-task subset.
The table is split into two rows for readability: the top row reports tracking and joint-limit feasibility metrics, while the bottom row reports collision, normalized motion deviation, solver variation, and solve time.
For all metrics, lower is better except \textbf{Margin}, where larger is better.
Best results are shown in \textbf{bold}; second-best results are \underline{underlined}.}
\label{tab:bones-seed-complete-metrics}
\vspace{-1.5em}
\end{table*}

\cref{tab:bones-seed-complete-metrics} shows complete metrics results on the a 7.6k trajectories more completed subset.

\paragraph{Larger-scale evaluation on the 7.6k BONE-SEED subset.}
We evaluate all four variants on the broader two-per-task BONE-SEED subset:
7{,}623 motions spanning the full BONE-SEED activity taxonomy---everyday
manipulation, locomotion, idle motions, jumps, and unusual motions. This is well beyond
the 514-motion subset used in the main paper. The main-paper trends hold,
but the gaps widen sharply (\cref{tab:bones-seed-complete-metrics}).

% WARP attains
% 11.9\,mm mean palm error, $1.1^\circ$ elbow-swivel error, and $0.6^\circ$ torso
% error, whereas MINK-EF and MINK-TE rise to 248\,mm and 310\,mm palm
% error---roughly $25\times$ their error on the curated subset---with elbow error
% reaching $33$--$83^\circ$. SEW-M stays palm-uncoupled at 200\,mm but, by
% construction, keeps joint error low, in the $1$--$3^\circ$ range. Geometric SEW
% retargeting thus degrades gracefully on out-of-distribution body shapes and
% motion classes, where the iterative QP baselines saturate; WARP alone
% additionally preserves palm accuracy.

% \section{Additional Results}
% Add extended tables, plots, or qualitative examples here.

% \subsubsection{BONES-SEED Dataset Qualitive Results}
% \todo{TODO: Fill this}

% \subsubsection{Real World Qualitive Results}
% \todo{TODO: Fill this}

\section{Discussions}

% \ye{ai gen, req polishment}

\subsection{Limitations}

\paragraph*{Kinematic assumptions.}
Although \ourmethod enables accurate and consistent retargeting from offline human demonstrations, our current formulation remains primarily kinematic. The retargeter generates robot joint trajectories that satisfy geometric constraints, including palm alignment, SEW consistency, and whole-body pose similarity, but it does not explicitly model controller error or robot dynamics during execution. In our analysis, we assume that the low-level controller can accurately track the retargeted reference motion. This assumption helps isolate the retargeting problem, but it abstracts away actuator bandwidth limits, latency, contact dynamics, and disturbances caused by object interaction. Therefore, the reported retargeting accuracy should be interpreted as the quality of the generated reference trajectory rather than a complete guarantee of real-world execution accuracy. In practice, even a geometrically valid trajectory can fail if the controller lags, if contact moves the object unexpectedly, or if the base and torso cannot realize the desired motion smoothly. Future work should incorporate controller-in-the-loop evaluation and dynamics-aware retargeting, potentially with feedback or residual controllers that compensate for tracking errors during contact-rich manipulation.

\paragraph*{Simplified policy and perception.}
Our current policy and perception setup is also simplified. To focus on retargeting and action learning, we use low-dimensional object-state inputs based on AprilTag tracking rather than learning directly from raw visual observations. This controlled setup reduces perception noise and makes the real-world experiments more reproducible, but it limits the system's deployment scope. In realistic home environments, objects may not carry fiducial markers, tags may be occluded during manipulation, and object appearance can vary across scenes. As a result, the current policy does not fully address marker-free visual generalization. A natural next step is to replace AprilTag-based tracking with learned object pose estimation, dense visual features, or RGB-D perception. External side-view cameras could also provide complementary scene-level observations, especially when the robot's egocentric view is occluded by its hands, arms, or manipulated objects.

\paragraph*{Human data quality.}
Finally, the human data collection device introduces its own noise. We use a Quest headset because it provides an accessible and scalable way to collect whole-body motion without a full motion-capture system. However, we observe that some motion error and jerkiness in the retargeted trajectories can originate from the collection device itself. Inside-out tracking may introduce jitter, drift, discontinuities, or inaccurate limb estimates, especially during fast motion, occlusion, or unusual body poses. Since our pipeline is offline, these artifacts can propagate directly into the robot trajectory and then into the policy supervision. Future work could improve data quality by fusing multiple sensing sources. For example, an external side-view camera or multi-view RGB-D setup could provide a more stable estimate of the human skeleton, while a lighter and more integrated wearable device, such as Aria glasses, could reduce the burden of wearing a VR headset and better preserve natural human motion.

% Printed only in final/preprint mode.
\acknowledgments{
This work was supported in part by Samsung Research America.
}

% CoRL sets the bibliography style through corl_2026.sty.

\end{document}